\documentclass[conference]{IEEEtran}
\usepackage{times}

\usepackage[numbers]{natbib}
\usepackage{multicol}

\usepackage[bookmarks=true, hidelinks]{hyperref}

\usepackage{amsmath} 
\usepackage{graphicx}
\usepackage{multirow}
\usepackage{booktabs}
\usepackage{amssymb}

\let\labelindent\relax
\usepackage{enumitem}
\usepackage{algorithm}
\usepackage{subcaption}
\usepackage[noend,]{algpseudocode}
\usepackage[export]{adjustbox}

\algrenewcommand{\algorithmiccomment}[1]{\hfill {\footnotesize\textcolor{blue}{$\triangleright$ #1}}}

\algnewcommand{\LineComment}[1]{{\\ \footnotesize\textcolor{blue}{$\triangleright$ #1}}}

\usepackage[table]{xcolor}

\usepackage{xcolor}
\usepackage{listings}

\usepackage{listings}
\definecolor{commentcolor}{rgb}{0.25,0.50,0.37}  
\definecolor{keywordcolor}{rgb}{0.13,0.13,1}     
\definecolor{stringcolor}{rgb}{0.63,0.13,0.94}   
\definecolor{emphcolor}{rgb}{1,0.5,0}            
\definecolor{mutedredorange}{rgb}{0.8, 0.4, 0.3}  

\lstdefinelanguage{PDDL}{
  alsoletter={:,-,=},
  morekeywords={:action, :parameters, :domain, :precondition, :stream, :output, :certified, :effect, :effects, :ueffects, :conditions, :uconds, :axiom, :condition, :reward, not, and, possibly, or, imply, exists, forall, increase, when},
  morecomment=[l]{;},
  morestring=[b]",
}

\lstdefinestyle{PDDLStyle}{
  language=PDDL,
  basicstyle=\ttfamily\small, 
  breaklines=true,
  columns=fullflexible,
  keywordstyle=\color{keywordcolor}\bfseries,
  commentstyle=\color{commentcolor},
  stringstyle=\color{stringcolor},
  emph={maybe, not, and, or, imply, exists, forall, increase, when},
  emphstyle=\bfseries,
  showstringspaces=false, 
}

\newcommand\restr[2]{{
  \left.\kern-\nulldelimiterspace 
  #1 
  \vphantom{\big|} 
  \right|_{#2} 
  }}
\newcommand\doubleplus{+\kern-1.3ex+\kern0.8ex}

\usepackage{todonotes}

\newif\ifshowpddl
\showpddlfalse

\newcommand{\showpddl}[1]{\ifshowpddl #1\fi}

\title{\LARGE \bf
Partially Observable Task and Motion Planning with \\Uncertainty and Risk Awareness
}

\author{
\textbf{Aidan Curtis,} \textbf{George Matheos,} \textbf{Nishad Gothoskar,} \textbf{Vikash Mansinghka,} \\
\textbf{Joshua Tenenbaum,} \textbf{Tomás Lozano-Pérez,} \textbf{Leslie Pack Kaelbling} \vspace{7px}
\\ 
MIT Computer Science and Artificial Intelligence Laboratory \\
{\tt\small \{curtisa, gmatheos, nishadg, vkm, tlp, lpk\}@mit.edu}.
}
\begin{document}
\maketitle

\thispagestyle{empty}
\pagestyle{empty}

\begin{abstract}
Integrated task and motion planning (TAMP) has proven to be a valuable approach to generalizable long-horizon robotic manipulation and navigation problems. However, the typical TAMP problem formulation assumes full observability and deterministic action effects. These assumptions limit the ability of the planner to gather information and make decisions that are risk-aware. We propose a strategy for TAMP with Uncertainty and Risk Awareness (TAMPURA) that is capable of efficiently solving long-horizon planning problems with initial-state and action outcome uncertainty, including problems that require information gathering and avoiding undesirable and irreversible outcomes. Our planner reasons under uncertainty at both the abstract task level and continuous controller level. Given a set of closed-loop goal-conditioned controllers operating in the primitive action space and a description of their preconditions and potential capabilities, we learn a high-level abstraction that can be solved efficiently and then refined to continuous actions for execution. We demonstrate our approach on several robotics problems where uncertainty is a crucial factor and show that reasoning under uncertainty in these problems outperforms previously proposed determinized planning, direct search, and reinforcement learning strategies. Lastly, we demonstrate our planner on two real-world robotics problems using recent advancements in probabilistic perception.

\end{abstract}
\IEEEpeerreviewmaketitle


\section{Introduction}

In an open-world setting, a robot's knowledge of the environment and its dynamics is inherently limited. If the robot believes it has full knowledge of the state and dynamics of the world, it may confidently take actions that have potentially catastrophic effects, and it will never have a reason to seek out information. For these reasons, it is crucial for the robot to know what it does not know and to make decisions with an awareness of risk and uncertainty.

\begin{figure}[ht]
    \centering
    
    \begin{subfigure}[b]{\linewidth}
        \centering
        \includegraphics[width=\linewidth]{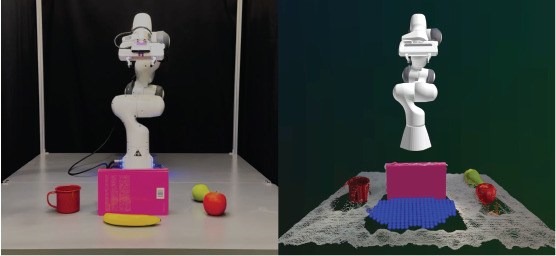}
    \end{subfigure}

   \vspace{0.2cm} 
   
    \begin{subfigure}[b]{\linewidth}
        \centering
        \includegraphics[width=\linewidth]{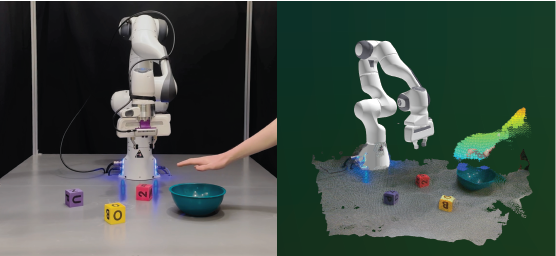} 
    \end{subfigure}
    \caption{Top: Robot with wrist mounted camera looking for a banana. The robot plans to take information gathering actions based on a posterior estimate of the banana's pose shown in blue. Bottom: Robot with one wrist mounted camera and one external camera plans to complete a long-horizon manipulation task while avoiding a human in the workspace.}
    \label{fig:teaser}
    \vspace{-6mm}
\end{figure}

\begin{figure*}[htbp]
    
    \centering
    \begin{subfigure}{0.18\textwidth}
        \includegraphics[width=\linewidth]{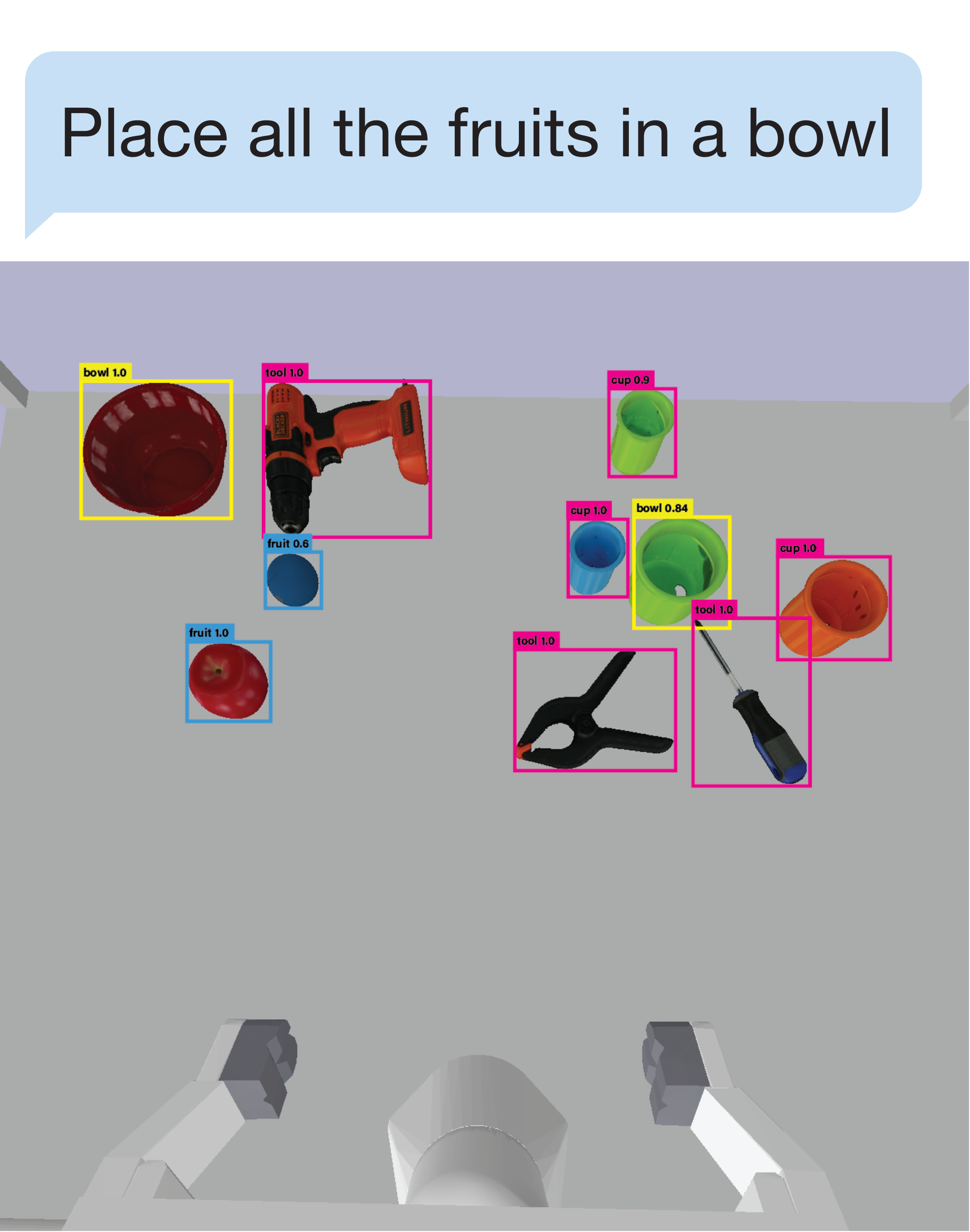}
        \caption{Class uncertainty}
    \end{subfigure}%
    \hfill
    \begin{subfigure}{0.18\textwidth}
        \includegraphics[width=\linewidth]{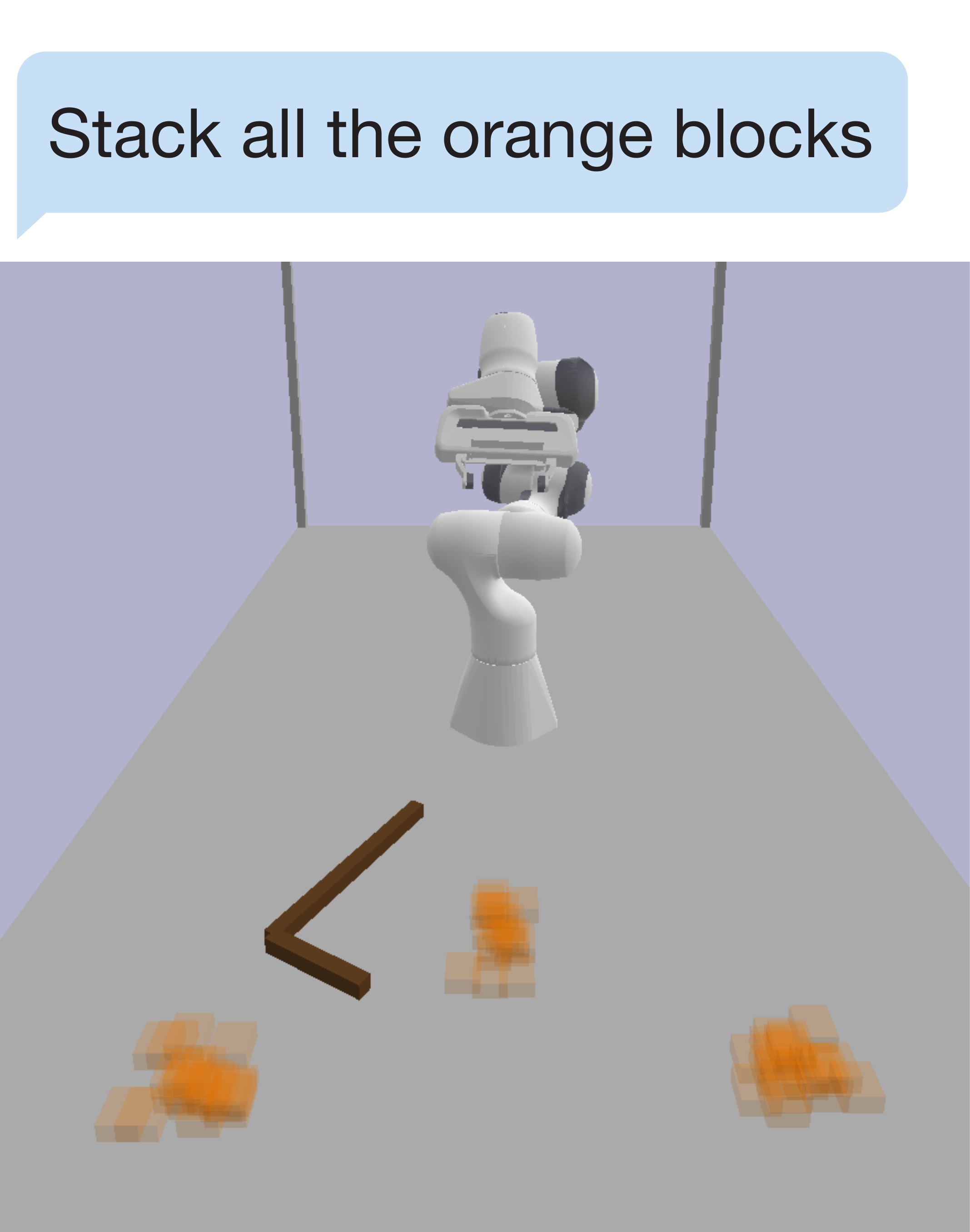}
        \caption{Pose uncertainty}
    \end{subfigure}%
    \hfill
    \begin{subfigure}{0.18\textwidth}
        \includegraphics[width=\linewidth]{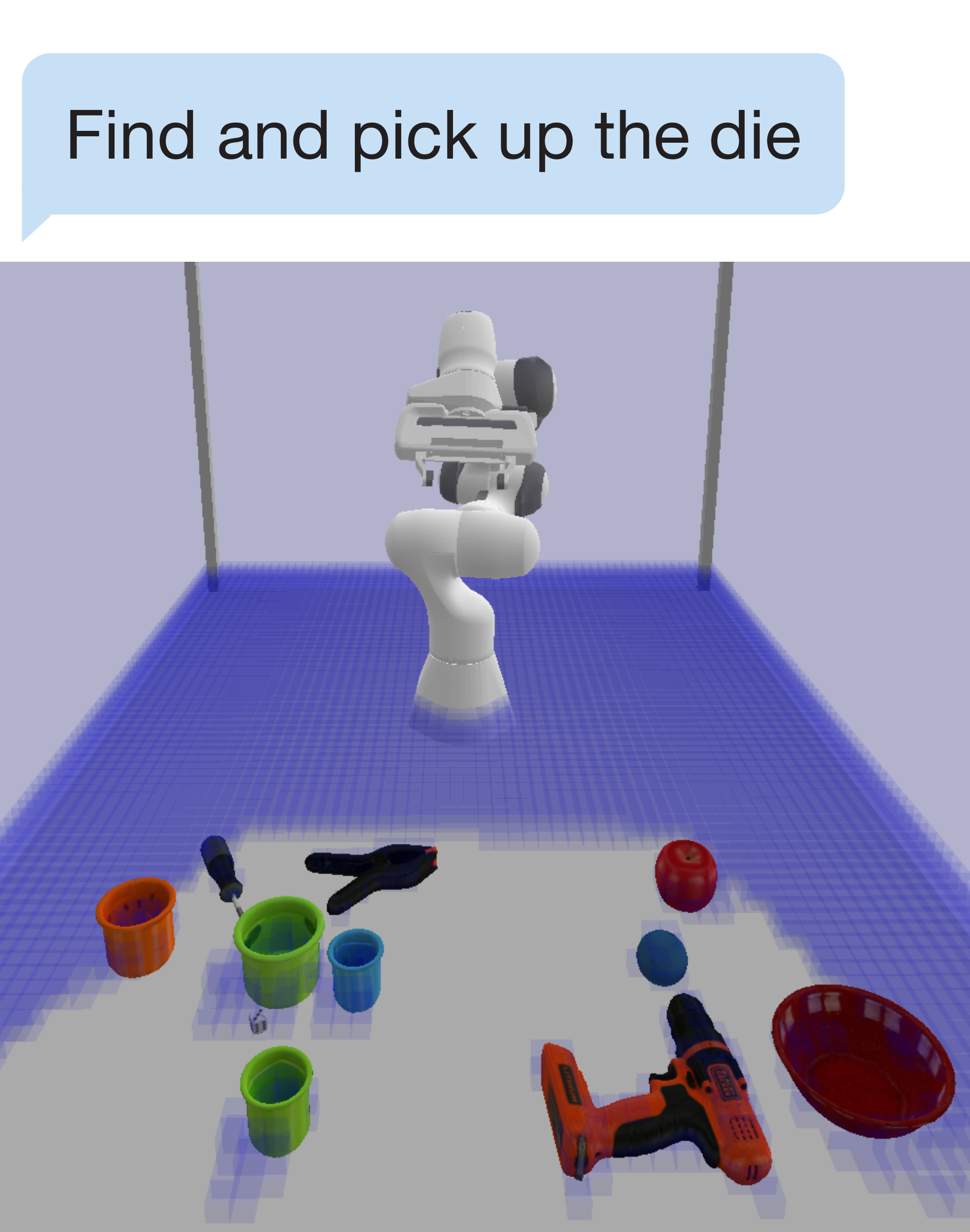}
        \caption{Partial observability}
    \end{subfigure}%
    \hfill
    \begin{subfigure}{0.18\textwidth}
        \includegraphics[width=\linewidth]{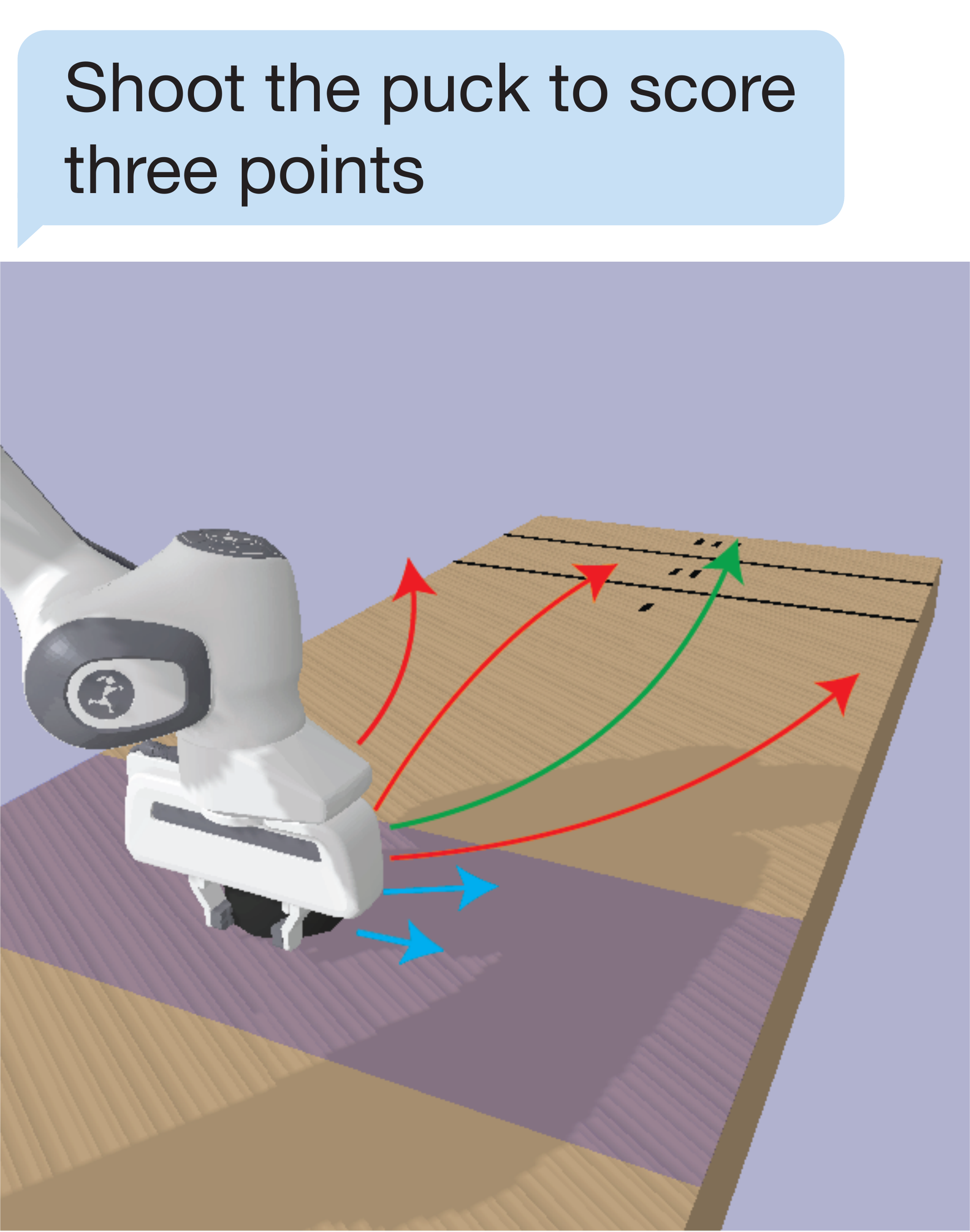}
        \caption{Physical uncertainty}
    \end{subfigure}%
    \hfill
    \begin{subfigure}{0.18\textwidth}
        \includegraphics[width=\linewidth]{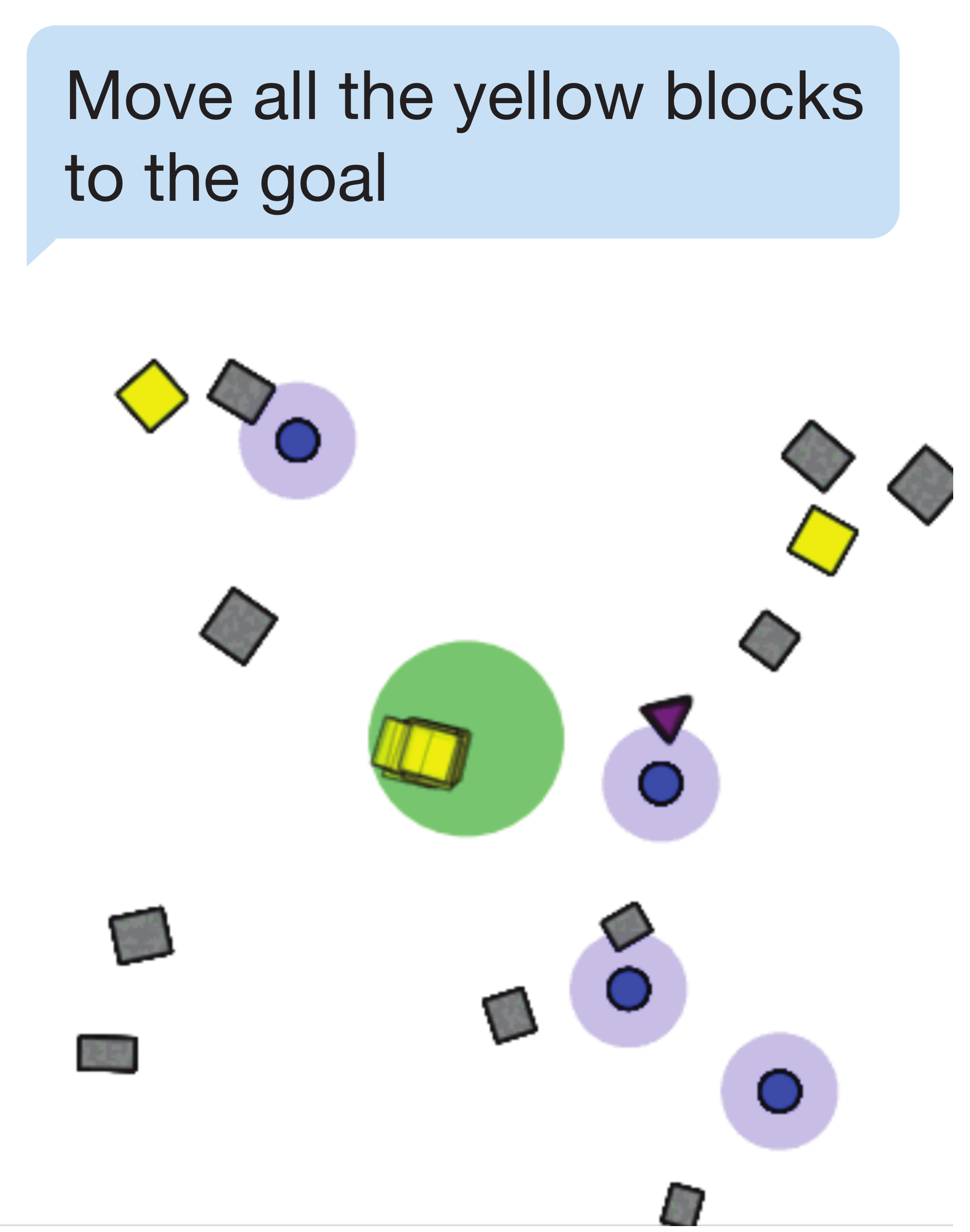}
        \caption{SLAM uncertainty}
    \end{subfigure}
    
    \caption{
    This figure illustrates five long-horizon planning tasks that TAMPURA is capable of solving. 
    Each of them contains a unique type of uncertainty including uncertainty in (a) classsification, (b) pose due to noisy sensors or (c) partial observability, (d) physical properties such as friction or mass, and (e) localization/mapping due to odometry errors}
    \label{fig:tasks}
    \vspace{-4mm}
\end{figure*}

Advances in techniques like behavior cloning (BC)~\cite{behavior_cloning, behavior_cloning_2}, reinforcement learning (RL)~\cite{robust_rl, mbrl_driving}, and model-based control~\cite{pang2023global, Wirnshofer2020ControllingCM} have made it possible to develop robotic controllers for short time-horizon manipulation tasks in partially observable or stochastic domains. 
In situations matching a narrow training distribution (BC), with dense reward and short horizons (RL), or conforming to modeling assumptions, these controllers can be quite robust. 
However, these methods typically do not generalize to solving arbitrary complex goals over long time horizons.

Simultaneously, recent advances in Task and Motion Planning (TAMP) have illustrated the viability of using planners to sequence such controllers to robustly achieve tasks over longer time horizons in large open-world settings~\cite{m0m, kumar2023learning, yang2023sequencebased}. 
In TAMP, the key to tractable planning over long time horizons is to sequence short-horizon controllers, exploiting a description of the conditions in which each controller can be expected to work, and of each controller's effects. 
However, most TAMP formulations assume that these symbolic descriptions perfectly and deterministically characterize the effects of running the controllers. 
In real robotics settings, stochasticity and partial observability make it impossible to exactly predict the effects of controllers. 
Furthermore, it is typically impossible to obtain exact symbolic descriptions that fully capture the effects and preconditions of each controller.

\begin{figure*}[t]
    \centering
    \includegraphics[width=0.85\linewidth]{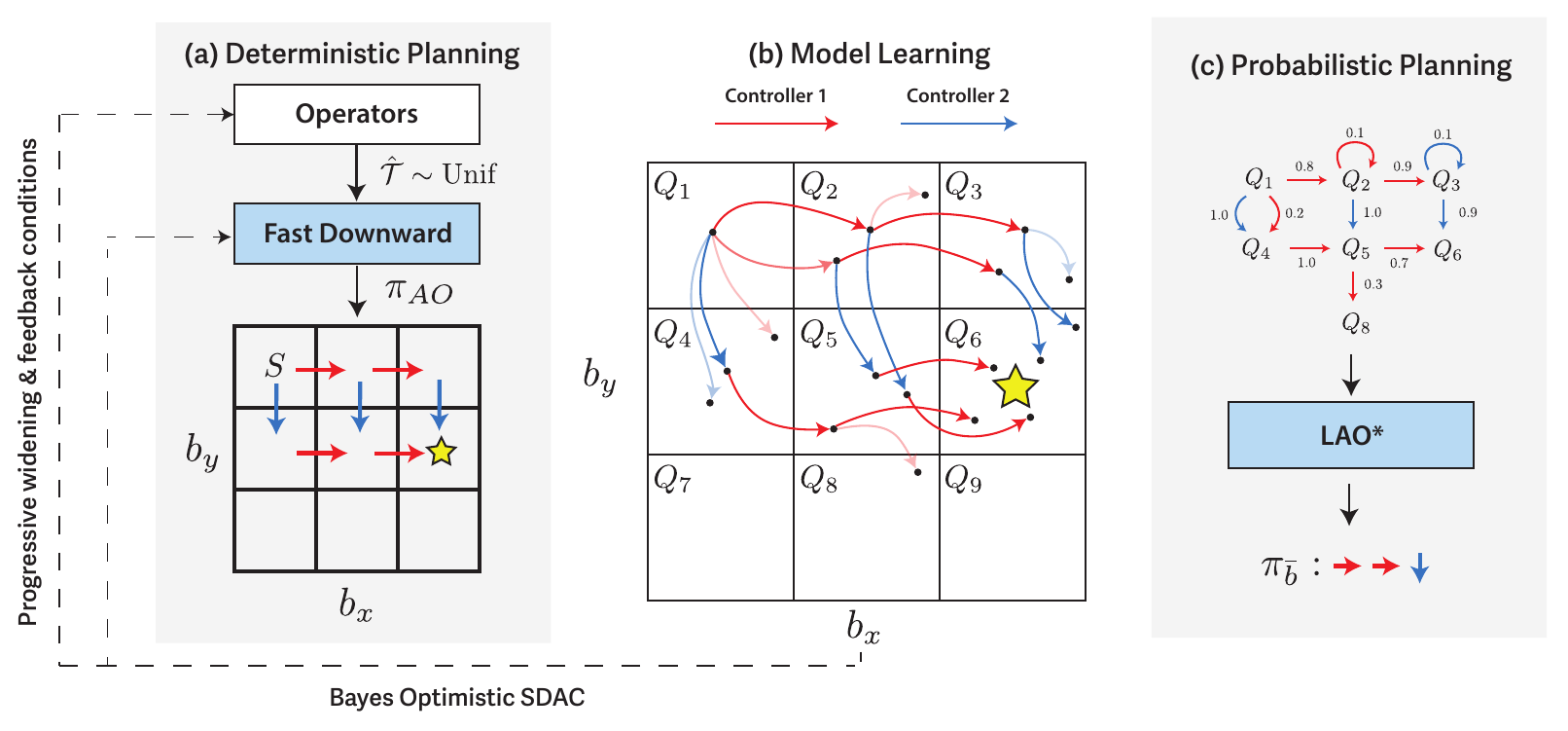}
    \caption{Uncertainty and Risk Aware Task and Motion Planning.  
    (a) The robot’s continuous space of probabilistic beliefs about world state is partitioned into a discrete abstract belief space, here with 9 states.
    TAMPURA considers a set of operators, each containing a low-level robot controller, and a description of the possible effects of executing the controller.
    Determinized planning computes possible sequences of controllers to reach the goal.  
    These plans do not factor in uncertainty or risk and would be unsafe or inefficient to execute in the real world.  
    (b) The determinized plans are executed in a mental simulation. The distribution of effects is recorded, to learn an MDP on the space of abstract belief states visited in these executions.  
    By iterating between determinized planning and plan simulation (Sec.~\ref{sec:tampura_algorithm}), TAMPURA learns a sparse MDP related to the original decision problem.
    (c) The robot calculates an uncertainty and risk aware plan in the sparse MDP, and executes it.}
    \label{fig:diagram}
    \vspace{-4mm}
\end{figure*}

This paper shows how to extend TAMP to settings with partial observability, uncertainty, and imperfect symbolic descriptions of controllers.  
Our approach, TAMPURA, is to exploit a coarse model of each controller's preconditions and effects to rapidly solve deterministic, symbolic planning problems that guide the construction of a non-deterministic Markov Decision Process (MDP) with only a small number of actions applicable from each state (Figure~\ref{fig:diagram}). 
This smaller model captures the key tradeoffs between utility, risk, and information-gathering in the original planning problem.  
The resulting MDP is sparse enough that high-quality uncertainty-aware solvers like LAO*~\cite{lao} can be applied. 
The guidance used to distill this small MDP comes in the form of symbolic descriptions of the preconditions, and the uncertain but possible effects, of each controller.
The MDP is constructed by learning the probability distribution over these possible effects for each controller, refining coarse and imperfect descriptions into transition distributions which can be used for uncertainty and risk-aware planning.
In this paper, we use controller descriptions provided by engineers; in Section~\ref{sec:discussion}, we comment on how future work could enable such descriptions to be learned or generated with large language models, following~\cite{loft} or~\cite{worms2worms}.

We demonstrate the applicability of TAMPURA in a wide range of simulated problems (Figure~\ref{fig:tasks}), and show how it can be applied to two real-world robotics tasks: searching a cluttered environment to find objects, and operating safely with an unpredictable human in the workspace (Figure~\ref{fig:teaser}).
We show that in tasks requiring risk sensitivity, information gathering, and robustness to uncertainty, TAMPURA significantly outperforms reinforcement learning, Monte Carlo tree search, and determinized belief-space task and motion planners, even when these algorithms are all given access to the same controllers.

\section{Related Work}

Planning under environment uncertainty and partial observability is a longstanding problem with a diversity of approaches. 
While exact methods are typically only suited for small discrete problems, approximate methods~\cite{pomcp, despot} have shown that online planning with frequent replanning can work well for many non-deterministic, partially observable domains with large state and observation spaces. 
These methods directly search within a primitive action space and are guided by reward feedback from the environment. 
However, in the absence of dense reward feedback, the computational complexity of these methods scales exponentially with the planning horizon, action space, and observation space.

Task and motion planning refers to a family of methods that solve long-horizon problems with sparse reward through temporal abstraction, factored action models, and primitive controller design~\cite{tamp_survey, pddlstream, m0m}.
While most TAMP solutions assume deterministic transition dynamics and full observability, several approaches have extended the framework to handle stochastic environments or partial observability. 
Some TAMP solvers remove the assumption of deterministic environments~\cite{stochastic_tamp, contingent_tamp, loft}. While these approaches can find contingent task and motion plans~\cite{stochastic_tamp} or use probabilities to find likely successful open-loop plans~\cite{loft}, they operate in state space and assume prior information about transition probabilities in the form of hardcoded probability values~\cite{stochastic_tamp}, or demonstration data in the form of example plans~\cite{loft}. 
In contrast, our proposed approach learns belief-space transition probabilities through exploration during planning.

Another family of TAMP solvers plan in belief space~\cite{SSReplan, bhpn, rohan_belief_tamp, contingent_tamp}, allowing them to plan to gather necessary information, even in long-horizon contexts. 
Some approaches, such as IBSP and BHPN do forward or backward search in the continuous belief space, respectively~\cite{rohan_belief_tamp, bhpn}. 
SSReplan~\cite{SSReplan} embeds the belief into the high-level symbolic model. 
While these approaches to TAMP in belief space have their tradeoffs, all of them perform some form of determinization when planning. 
Determinization allows the planner to only consider one possible effect of an action, which inherently limits its ability to be risk-aware, incorporate action cost metrics, and perform well under large observational branching factors. 

To our knowledge, there exists one other belief space TAMP planner that does not determinize the transition dynamics during planning~\cite{contingent_tamp}. 
However, this approach focuses on contingent planning, where there are no probabilities associated with nondeterministic outcomes. 
In our experiments, we compare to many of these approaches to show the power of stochastic belief-space planning with learned action effect probabilities.

\section{Background}
\label{sec:background}
One way to formulate many sequential decision problems involving uncertainty is as Partially Observable Markov Decision Processes (POMDPs).  
A POMDP is a tuple $\mathcal{M} = \langle \mathcal{S}, \mathcal{O}, \mathcal{A}, \mathcal{T}, \mathcal{Z}, r, b_0, \gamma \rangle$.\footnote{A reference for all notation introduced henceforth is provided in Table~\ref{table:nomenclature} in the appendix.}
$\mathcal{S}, \mathcal{O},$ and $\mathcal{A}$ are the state, observation, and action spaces.
The state transition and observation probabilities are $\mathcal{T}(s_{t+1} \mid s_t, a_t)$ and $\mathcal{Z}(o_t \mid s_t)$, and $b_0$ is a probability distribution on $\mathcal{S}$ giving the distribution of possible initial states.  
The reward function is $r(s_t, a_t, s_{t+1})$, and $\gamma$ is a discount factor quantifying the trade-off between immediate and future rewards.

\subsection{Belief-State MDP} 
\label{sec:belief_state_mdp}
From any POMDP, one can derive the continuous belief-space MDP $\mathcal{M}_b = \langle \mathcal{B}, \mathcal{A}, \mathcal{T}_b, r_b, b_0, \gamma \rangle$~\cite{pomdps_are_mdps}. 
The state space of this MDP is $\mathcal{B}$, the space of probability distributions over $\mathcal{S}$, or belief states.  
The initial belief state is $b_0 \in \mathcal{B}$, describing the robot's belief before any actions have been taken or any observations have been received.
The reward $r_b$ is derived from $r$; $\gamma$ is unchanged.
The transition distribution
$\mathcal{T}_b(b_{t+1} \mid b_t, a_t)$ is the probability that after being in belief state $b_t$ and taking action $a_t$, the robot will receive an observation causing it to update its belief to $b_{t+1}$.

\subsection{Belief updates}
\label{sec:belief_updates}
As TAMPURA plans in a belief-space MDP, it must keep track of the robot's current belief state.
However, computing the belief updates exactly is intractable in many problems.
Fortunately, in cases where exact belief updates cannot be computed, it can suffice to compute approximate belief states using approximate Bayesian inference methods like particle filtering, or more generally, sequential Monte Carlo~\cite{chopin_smc, probroboticsbook}. 
Further, recent advances in probabilistic programming~\cite{gen, gensp, smcp3, imcmc} and its application to 3D perception~\cite{gothoskar2023bayes3d, 3dp3, nel} have made it practical to generate belief distributions over latent states describing the poses of 3D objects, their contact relationships, and to update these beliefs in light of new RGB-Depth images of a 3D scene. 
In our real-world robotic experiments, probabilistic perception is performed using Bayes3D~\cite{gothoskar2023bayes3d}.

\subsection{Belief-State Controller MDP}
\label{sec:belief_state_controller_mdp}
When the action space $\mathcal{A}$ represents primitive controls to the robot such as joint torques or end-effector velocity commands, the time horizons to perform meaningful tasks can be enormous, rendering planning intractable. To mitigate this, we introduce the concept of a belief-space controller, which takes the current belief as input and executes in closed-loop fashion over extended time horizons.
For example, in our 2D SLAM task modeling a mobile robot moving with pose uncertainty, our ``move to point X'' controller has access to a distribution over possible robot poses.
The controller selects actions that would not result in collisions under any possible poses in the support of this distribution, sometimes ruling out unsafe low-level actions that would seem safe and more efficient given only the most likely pose of the robot.

Given a set of learned or designed controllers, the primitive belief-space MDP can be lifted to a temporally abstracted belief-space controller MDP $\mathcal{M}_c = \langle \mathcal{B}, \mathcal{C}, \mathcal{T}_c, r_c, \gamma \rangle$.  
The action space of this MDP is the space of controllers, and the transition model $\mathcal{T}_c(b_{t + 1} \mid b_t, c)$ gives the probability that if controller $c \in \mathcal{C}$ is executed beginning in state $b_t$, after it finishes executing, the belief state will be $b_{t+1}$. 
This paper considers the problem of planning in this belief-state controller MDP, to enable a robot to determine which low-level controllers to execute at each moment.  

\section{Planning with an abstract belief-state MDP}
\label{sec:abstraction}
Direct search in $\mathcal{M}_c$ is intractable in many problems due to the large branching factors in the action space and continuous belief outcomes.
To make the problem tractable,
TAMPURA applies a series of reductions to $\mathcal{M}_c$, ultimately producing a sparse abstract MDP $\mathcal{M}_s$ that can be solved efficiently with a probabilistic planner (Figure~\ref{fig:diagram}).

The reduction from $\mathcal{M}_c$ to $\mathcal{M}_s$ is performed in several steps.  
First, we perform belief-state abstraction, lifting from an MDP on $\mathcal{B}$ to an abstract MDP on $\overline{\mathcal{B}}$, which is a partition of $\mathcal{B}$ that groups operationally similar beliefs (Section~\ref{sec:abstract_bs_mdp}).  
Second, we leverage symbolic information describing the preconditions and possible but uncertain effects of each controller $c \in \mathcal{C}$ (Section~\ref{sec:operators}) to construct a determinized shortest path problem on the abstract belief space. 
Determinized planning in abstract belief space with controller descriptions is now tractable, but the resulting plans are not risk-aware due to determinization. 
In addition, the resulting plans may be overly optimistic because they are untethered from geometric and physical constraints. 
Third, we approximately learn the transition model $\overline{\mathcal{T}}$ on $\overline{\mathcal{B}}$, using the efficient determinized planner to focus exploration on the task-relevant parts of the abstract belief space (Section~\ref{sec:model_learning}).  The subset $\mathcal{B}_\text{sparse} \subseteq \overline{\mathcal{B}}$ of abstract belief states focused on in model learning forms the state space of the sparse abstract MDP; its transition model is the learned transition probabilities, $\hat{\mathcal{T}}$, approximating the true $\overline{\mathcal{T}}$.
This sparse MDP distills key tradeoffs about risk, information-gathering, and outcome uncertainty from the original problem. It can be solved with a probabilistic planner such as LAO*, resulting in a risk and uncertainty aware policy in the abstract belief-state MDP. The first action recommended by this policy is the next controller to execute on the robot. The full TAMPURA robot control loop is given in Algorithm~\ref{alg:tampura}.

\subsection{Belief state propositions}
\label{sec:belief_state_propositions}
To apply TAMPURA in a controller-level MDP $\mathcal{M}_c$, an abstract belief space must be defined through the specification of a set $\Psi_\mathcal{B}$ of belief state propositions.  Each $\psi \in \Psi_\mathcal{B}$ is a boolean function $\psi : \mathcal{B} \to \{0, 1\}$ of the robot's belief.
As described in~\citep{bhpn}, belief propositions can be used to describe comparative relationships such as \texttt{(MLCat ?o ?c)}, meaning the most likely category of object $\texttt{?o}$ is category $\texttt{?c}$, statements about the probable values of object properties such as \texttt{(BPose ?o ?x)}, meaning the probability object $\texttt{?o}$'s position is within $\delta$ of $\texttt{?x}$ is greater than $1 - \epsilon$, and other statements about the distribution over a value such as \texttt{(BVPose ?o)}, meaning there exists some position $\texttt{?x}$ such that object $\texttt{?o}$ is within $\delta$ of $\texttt{x}$ with probability greater than $1 - \epsilon$. In practice, these are implemented as lifted symbol grounding functions that take in a set of entities along with the current belief and return a boolean value.

\subsection{The abstract belief-state MDP}
\label{sec:abstract_bs_mdp}
Given a particular belief $b\in\mathcal{B}$, evaluation under the proposition set results in an abstract belief $\text{abs}(b) := \{\psi \mapsto \psi(b) : \psi \in \Psi_\mathcal{B}\}$ which is a dictionary mapping from each proposition $\psi$ to its value $\psi(b)$ in belief $b$. The abstract belief space is then defined as $\overline{\mathcal{B}} := \{\text{abs}(b) : b \in \mathcal{B}\}$.
Under a particular condition on this set of propositions (Appendix~\ref{app:stationarity}) which we assume to hold in this paper,\footnote{TAMPURA can be run when this condition does not hold and we expect its performance to degrade gracefully in this case. 
See Appendix~\ref{app:stationarity} for details.} we can derive from $\mathcal{M}_c$ an abstract belief-state controller MDP $\overline{\mathcal{M}_c} := \langle \overline{\mathcal{B}}, \mathcal{C}, \overline{\mathcal{T}}, r_c, \bar{b}_0, \gamma \rangle$.  
The state space of this MDP is the discrete abstract belief space $\overline{\mathcal{B}}$, rather than the uncountably infinite belief space $\mathcal{B}$; transition probabilities on $\overline{\mathcal{B}}$ are given by $\overline{\mathcal{T}}$. 
The initial state is $\bar{b}_0 := \text{abs}(b_0)$. 

In this paper, we focus on planning problems with objectives modeled as goals in belief space (e.g., the goal may be to believe that with high probability the world is in a desired set of states.)
Specifically, we consider the case where the reward is a subset of the abstract belief space: $G \subset \overline{\mathcal{B}}$ such that the goal can be defined in terms of the belief-space propositions.   
We also model episodes as terminating at the first moment a goal belief state is achieved.  This restricts the controller-level MDP under consideration to a belief-space stochastic shortest paths problem (BSSP).

\subsection{Operators with uncertain effects}
\label{sec:operators}

\setlength{\textfloatsep}{4pt}
\begin{algorithm}[t]
\caption{TAMPURA Control Loop}
\label{alg:tampura}
\begin{algorithmic}[1]
\Require Planning problem: $(\mathcal{T}_c, \mathbb{O}, r_c, b_0)$
\State $s \gets \emptyset$ \Comment{Initialize state used by model learning.}
\State $\mathcal{B}_\text{sparse}, \hat{\mathcal{T}} \gets \emptyset, \emptyset$ \Comment{Initialize result of model learning.} 
\While{$\text{abs}(b) \notin G$}
\If{$\text{abs}(b) \notin \mathcal{B}_\text{sparse}$}
\State $\text{args} \gets (b_0, G, \mathbb{O}, s)$
\State $s, \hat{\mathcal{T}}, \mathcal{B}_\text{sparse} \gets \texttt{Model-Learning}(\text{args})$ \label{alg:line:call_model_learning}
\EndIf
\LineComment{Solve the MDP with $\hat{\mathcal{T}}$ and $ r_c$ over $\mathcal{B}_\text{sparse}$}
\State $\pi \gets \texttt{LAO-Star}(\mathcal{B}_\text{sparse}, \hat{\mathcal{T}}, r_c)$
\LineComment{Get controller recommended by policy $\pi$ in $b_0$.}
\State $c \gets \pi(\text{abs}(b_0))$
\State $(\vec{o}, \vec{a}) \gets \texttt{Execute}(c)$
\State $b_0 \gets \texttt{BeliefUpdate}(b_0, \vec{o}, \vec{a})$
\EndWhile
\end{algorithmic}
\end{algorithm}

These belief-space propositions give us a language to describe the preconditions and possible effects of executing a controller. 
We call such descriptions operators $\texttt{op} \in \mathbb{O}$.
Each operator is a tuple $\langle \texttt{Pre}, \texttt{Eff}, \texttt{UEff}, \texttt{UCond}, c \rangle$.
Here, $\texttt{Pre} \subseteq \Psi_{\mathcal{B}}$ is the set of belief propositions that must hold for a controller $c \in \mathcal{C}$ to be executed, $\texttt{Eff} \subseteq \Psi_{\mathcal{B}}$ is the set of effects that are guaranteed to hold after $c$ has been executed, $\texttt{UEff} \subseteq \Psi_{\mathcal{B}}$ is the set of belief propositions that have an unknown value after the completion of $c$, and $\texttt{UCond} \subseteq \Psi_{\mathcal{B}}$ represents the set of propositions upon which the probability distribution over the $\texttt{UEff}$ may depend.
(That is, given an assignment to the propositions in $\texttt{UCond}$, there should be a fixed distribution on $\texttt{UEff}$, though this distribution need not be known a priori.)

As a result of this additional structure, from any given abstract belief space $\bar{b} \in \overline{\mathcal{B}}$, only a small number of operators can be applied, as most operators will not have their preconditions satisfied.  
Additionally, planners can exploit the knowledge that after applying an operator from state $\bar{b}$, the only reachable new states are those which modify $\bar{b}$ by turning on the propositions in $\texttt{Eff}$, and possibly turning on some propositions in $\texttt{UEff}$.

\subsection{Operator schemata} 
\label{sec:example_operator}

In our implementation, the set of operators and the set of controllers are generated from a set of operator schemata.  
Each operator schema describes an operation which can be applied for any collection of entities with a given type signature, for any assignment to a collection of continuous parameters the controller needs as input. 
$\mathbb{O}$ is the set of grounded, concrete operators generated from an assignment of objects and continuous parameters to an operator schema.  

We introduce an extension to PDDL for specifying schemata for controllers with uncertain effects.  
An example operator schema written in this PDDL extension is shown below.

\begin{lstlisting}[style=PDDLStyle]
(:action pick
 :parameters (?o - object ?g - grasp)
 :precondition (and (BVPose ?o) (BHandFree))
 :effects (and $\lnot$ (BVPose ?o))
 :uconds (and (BClass ?o @glass))
 :ueffects (maybe (Broken ?o) (BGrasp ?o ?g)))
\end{lstlisting}


For any entity $\texttt{o}$ with $\text{type}($\texttt{o}$) = \texttt{object}$, and any continuous parameter $\texttt{g}$ with type $\text{type}(\texttt{g}) = \texttt{grasp}$, this operator schema yields a concrete operator $\texttt{pick}_{o, g} \in \mathbb{O}$.  As specified in the \texttt{:precondition}, this operator can only be applied from beliefs where the pose of $o$ is known $\texttt{(BVPose ?o)}$ and the robot's hand is believed to be free $\texttt{(BHandFree)}$.  
As specified in \texttt{:effects}, after running this, it is guaranteed that there will not exist any pose $p$ on the table such that the robot believes $o$ is at $p$ with high probability.
The \texttt{:ueffects} field specifies two possible but not guaranteed effects. Following a controller execution, these effects evaluated on the updated belief belief using the symbol grounding functions in $\Psi_\mathcal{B}$. 
The overall effect of this operator can be described as a probability distribution on the four possible joint outcomes.
The \texttt{:uconds} field specifies that this probability distribution will be different when $\texttt{o}$ is believed to be glass than when it is not. 
Such a difference in outcome distributions may lead the planner to inspect the class of an object before attempting to grasp it.  Our semantics are similar to those in PPDDL~\cite{ppddl} and FOND~\cite{fond}, but are agnostic to the exact outcome probabilities and ways in which the conditions affect those probabilities.

\section{Learning the Sparse Abstract MDP}
\label{sec:model_learning}

While the operator schemata are helpful for guiding planning, they lack outcomes probabilities that are crucial for finding a kinematically and geometrically valid plan that is safe and efficient. 
\footnote{Transition probabilities can capture geometric and kinematic constraints that hard symbolic constraints do not rule out.  For instance, a controller $\texttt{pick}_{o, g}$ may have $\texttt{(BGrasp ?o ?g)}$ in its $\texttt{UEff}$s, even for a grasp $g$ which is kinematically infeasible.  This infeasibility will be captured once the transition probabilities are learned: the outcome has probability 0.
}
For any controller $c$ and any abstract belief state $\bar{b}$, it is possible to learn the outcome distribution $\overline{\mathcal{T}}(\cdot \mid \bar{b}, c)$ by simulating executions of $c$ from belief states consistent with $\bar{b}$. 
However, obtaining estimates of these probabilities is computationally expensive as it can involve geometric calculations, perceptual queries, and simulations.
Naive strategies like learning transition probabilities for $(\bar{b}, c)$ pairs sampled at random is highly inefficient (Figure~\ref{fig:mdp_experiments}, panel 2).
Our solution is to leverage the symbolic structure and specified goal to determine which outcome distributions to learn for more efficient online model learning.

\subsection{Solution-guided model learning}
\label{sec:solution_guided}
A seemingly natural strategy for goal-directed model learning is to first initialize $\overline{\mathcal{T}}$ so that $\overline{\mathcal{T}}(\cdot \mid \bar{b}_t, c)$ is uniform on the set of symbolically possible abstract belief states $\bar{b}_{t+1}$ specified in the $\texttt{UEffs}$ of the operator corresponding to controller $c$. 
After solving the abstract belief state MDP derived from the partially learned $\overline{\mathcal{T}}$ to obtain a policy $\pi$, we could simulate $\pi$, producing a sequence of $(\bar{b}_t, c_t, \bar{b}_{t+1})$ transitions. The transitions could be used to update the transition probabilities and construct a more accurate MDP in a process of iterative improvement.
The problem with this approach is that using the maximum likelihood estimate of the transition probabilities to guide exploration can converge to local optima, due to under-exploring actions for which the initial experience pool is poor.
Workarounds like $\epsilon$-greedy exploration can alleviate this, but are inefficient in problems with large action spaces as they explore the locally feasible action space rather than
focusing on task-relevant actions (Figure~\ref{fig:mdp_experiments}, panel 5).  
For example, in a setting where the robot must pick up a particular object, local exploration would experiment with picking unrelated objects.

\setlength{\textfloatsep}{4pt}
\begin{algorithm}[t!]
\caption{TAMPURA Online Model Learning} \label{alg:model_learning}
\begin{algorithmic}[1]
\Require Parameters for Bayesian model learning prior: $\alpha, \beta$
\Require Parameters controlling runtime: $I, K, S$
\Require Planning problem: $(b_0, G, \mathbb{O})$
\Require State from past iterations of model learning: $s$

\If{$s = \emptyset$} 
\label{line:start_counts}
\LineComment{Initialize count dictionaries w/ default value 0.}
\State $N \gets \texttt{DefaultDict}(\{\}, \text{default}=0)$
\State $D \gets \texttt{DefaultDict}(\{\}, \text{default}=0)$

\LineComment{Initialize dict from abstract beliefs to corresponding concrete beliefs.}
\State $P_\mathcal{B} \gets \texttt{DefaultDict}(\{\text{abs}(b_0): [b_0]\}, \text{default}=[])$
\Else
\State $(N, D, P_\mathcal{B}) \gets s$
\EndIf
\label{line:end_counts}
\LineComment{Main model learning loop.}
\For{$i = 1, \dots, I$} 

\LineComment{Plan + concatenate + filter $K$ trajectories.}
\State $(\tau_k)_{k=1}^K \gets \text{Determinized-Planner}(\bar{b}_0, K, \mathbb{O}, N, D, G)$ \label{line:plan}
\State $\tau^* \gets [(\bar{b}, \texttt{op}, \bar{b}') \in \text{concat}(\tau_1, \dots, \tau_k) : P_\mathcal{B}[\bar{b}]\neq []]$ \label{line:filter}
\LineComment{Compute preconditions + effects for each transition.}
\State $\vec{\Psi}_{\text{pre}} \gets [[\bar{b}[\psi] : \psi \in \texttt{op}.\texttt{UCond}] : (\bar{b}, \texttt{op}, \bar{b}') \in \tau^*]$
\State $\vec{\Psi}_{\text{eff}} \gets [[\bar{b}'[\psi] : \psi \in \texttt{op}.\texttt{UEffs}] : (\bar{b}, \texttt{op}, \bar{b}') \in \tau^*]$
\LineComment{List of controllers.}
\State $\vec{c} \gets [\texttt{op}.c : (\bar{b}, \texttt{op}, \bar{b}') \in \tau^*]$

\LineComment{Look up $s$: num times $c$ with preconditions $\Psi_\text{pre}$ led to
effects $\Psi_\text{eff}$.
}
\State $\vec{s} \gets [D[x] : x \in \text{zip}(\vec{\Psi}_{\text{pre}}, \vec{c}, \vec{\Psi}_{\text{eff}})]$
\LineComment{Compute $f$, num ``failures'' where $c$ in $\Psi_\text{pre}$ did not cause $\Psi_\text{eff}$.}
\State $\vec{f} \gets [
N[\Psi_{\text{pre}}, c] - s : (\Psi_{\text{pre}}, c, s) \in \text{zip}(\vec{\Psi}_{\text{pre}}, \vec{c}, \vec{s})
]$
\LineComment{Compute entropy $H$ to focus simulations on uncertain cases.}
\State $\vec{H} \gets [H(\alpha + s, \beta + f) : (s, f) \in \text{zip}(\vec{s}, \vec{f})]$ \label{line:entropy}

\LineComment{Controller simulation loop.}
\For{$j = 1, \dots, S$} 
\State $(\bar{b}_1, \texttt{op}, \bar{b}_2) \gets \texttt{pop}(\tau^*, \text{argmax}(\vec{H}))$\label{line:end_entropy}
\State $b_1 \sim \texttt{Unif}(P_\mathcal{B}[\bar{b}_1])$ \label{line:sample}
\State $b_2 \gets \texttt{Simulate}(b_1, \texttt{op}.c)$ \label{line:simulate}
\State $P_\mathcal{B}[\text{abs}(b_2)] \gets \texttt{Append}(P_\mathcal{B}[\text{abs}(b_2)], b_2)$
\State $\Psi_\text{pre} \gets [\bar{b}_1[\psi] : \psi \in \texttt{op}.\texttt{UCond}]$ \label{line:effects_abstraction_start}
\State $\Psi_\text{eff} \gets [\psi(b_2) : \psi \in \texttt{op}.\texttt{UEff}]$ \label{line:effects_abstraction_end}
\State $N[\Psi_\text{pre}, \texttt{op}.c] \gets N[\Psi_\text{pre}, \texttt{op}.c] + 1$
\State $D[\Psi_\text{pre}, \texttt{op}.c, \Psi_\text{eff}] \gets D[\Psi_\text{pre}, \texttt{op}.c, \Psi_\text{eff}] + 1$\label{line:update_counts_end}
\EndFor
\LineComment{Compile transition counts to sparse abstract MDP (Appendix~\ref{app:model_compilation}).}
\State $\hat{\mathcal{T}}, \mathcal{B}_\text{sparse} \gets \texttt{Compile}(D, N, \mathbb{O})$\label{line:compile}

\EndFor

\State \Return $(N, D, P_\mathcal{B}), \hat{\mathcal{T}}, \mathcal{B}_\text{sparse}$

\end{algorithmic}
\end{algorithm}

\subsection{Bayes optimistic model learning}
\label{sec:bayes_optimistic_model_learning}
Ideally, we would like our exploration strategy to be optimistic in the face of model uncertainty. One standard way of implementing optimism in a planning framework is with all-outcomes determinization~\cite{ffreplan}, wherein the planner is allowed to select the desired outcome of an action. This is done by augmenting the MDP action space with the possible action outcomes, resulting in a deterministic transition function \( \mathcal{T}_{AO} : \mathcal{B} \times (\mathcal{A}\times\mathcal{B}) \rightarrow \mathcal{B} \).

This optimism leads to bad policies when useful outcomes occur with low probability. To avoid this, cost weights $J$ can be added to the actions such that selecting outcomes with low probability is penalized. An optimal policy under the all-outcomes determinized model is an optimal open loop plan when cost weights are set to be $-\log(p)$ where $p$ is the true outcome probability~\cite{bhpn}. We make use of both of these strategies by initially collecting deterministic plans from a fully optimistic transition model, simulating the optimistic plans to gather transition data, and increasing the costs of outcomes as we gain certainty about the true transition probabilities. 

We model partial knowledge about each outcome probability $\overline{\mathcal{T}}(\bar{b}_{t+1} \mid \bar{b}_t, c_t)$ using a $\text{Beta}(\alpha, \beta)$ distribution. (For our prior we use $\alpha=1, \beta=1$.) Given simulations of $c_t$ from $\bar{b}_t$, the updated posterior is
$\text{Beta}(\alpha+s, \beta+f)$, where $s$ is the number of ``successful'' simulations which led to $\bar{b}_{t+1}$ and $f$ is the number of other ``failed'' simulations.

To model optimism in the face of uncertainty, we set outcome costs in all-outcomes determinized search according to an upper confidence bound of the estimated probabilities. Since we model outcome probabilities using beta distributions, we use a Bayesian-UCB~\cite{b-ucb} criterion where the upper bound is defined by the $\nu$ quantile of the posterior beta distribution. 
This quantile decreases with respect to the total number of samples across all sampled outcomes. 
As in any UCB, the rate of this decrease is a hyperparameter, but a common choice is $\nu=1/i$ because it leads to sublinear growth in risk and asymptotic optimality under Bernoulli distributed rewards. 
At the $i$th iteration of model learning, the Bayesian-UCB criterion corresponds to using all-outcomes costs
\begin{equation}
\label{equation:bayes_cost}
    J(\bar{b}_t, c_t, \bar{b}_{t+1}) = \log \big[ F_{\text{Beta}(\alpha + s, \beta + f)}^{-1}(1 - \frac{1}{i}) \big]
\end{equation}
Here, $J(\bar{b}_t, c_t, \bar{b}_{t+1})$ is the cost applied to transition $(\bar{b}_t, c_t) \to \bar{b}_{t+1}$ in all-outcomes planning, $F^{-1}$ is the inverse CDF of the Beta posterior, and $s$ and $f$ are as above.

\begin{figure*}[t!]
    \centering
    \captionsetup{aboveskip=-2pt}
    \includegraphics[width=\linewidth]{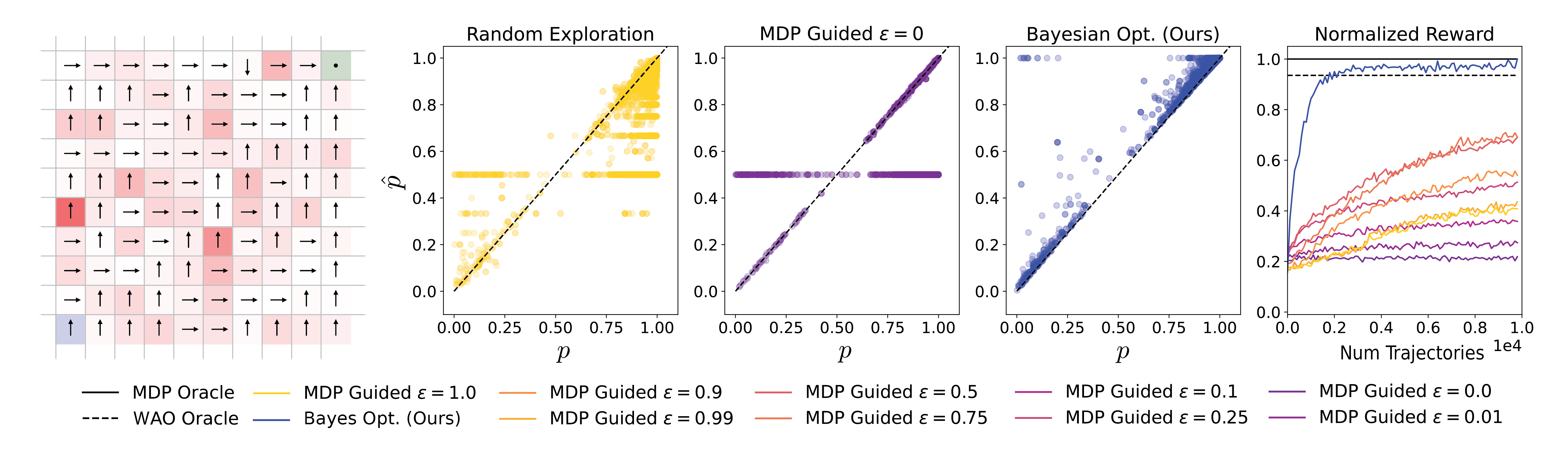}
    \caption{Comparisons of model-learning strategies on a simplified grid-world environment in which an agent must navigate from the blue cell to the green cell. Red intensity corresponds to $p$, the probability of transitioning to an irrecoverable state. $p$ for each cell is initially unknown, and must be estimated through interaction with the environment. The optimal policy given known $p$ for this sample environment is indicated with arrows. 
    The scatter plots compare the estimated $\hat{p}$ to true $p$ at the end of model learning for several strategies across 50 different environments.
    The rightmost plot shows average normalized reward as a function of the number of training trajectories for our method as well as the MDP-guided method with a variety of values of epsilon.
    Our method quickly reaches near optimal performance, surpassing the weighted all-outcomes determinized solution under ground truth outcome probabilities.}
    \label{fig:mdp_experiments}
    \vspace{-4mm}
\end{figure*}

This approach to model learning explores in the space of symbolically feasible goal directed policies, which is significantly more efficient than random action selection in problems with large action spaces and long horizons. 
In Figure~\ref{fig:mdp_experiments} we compare our Bayes optimistic model learning strategy to the solution guided strategy described in Section~\ref{sec:solution_guided}. Our experiments show that the bayes-optimistic approach to model learning outperforms $\epsilon$-greedy for all values of $\epsilon$ even in a domain with a relatively small action space. Note that although we use determinization to attain optimism in model learning, we perform full probabilistic planning on the learned model, making the resulting policy risk-aware.

\subsection{The TAMPURA model-learning algorithm}
\label{sec:tampura_algorithm}

In this section, we describe the details of the TAMPURA model learning algorithm outlined in Algorithm~\ref{alg:model_learning}.
For each task-relevant operator $\texttt{op} \in \mathbb{O}$, model learning must learn a probability table which induces a distribution over joint assignments to the propositions in $\texttt{op}.\texttt{UEffs}$, given any joint assignment to the propositions in $\texttt{op}.\texttt{UConds}$.
Algorithm~\ref{alg:model_learning} writes $\Psi_\text{pre}$ to denote assignments to an operator's $\texttt{UConds}$ and $\Psi_\text{eff}$ for assignments to $\texttt{UEffs}$.
This probability table is a compressed representation of $\overline{\mathcal{T}}$: for any symbolically feasible transition $(\bar{b}_t, c_t, \bar{b}_{t+1})$, the value of $\overline{\mathcal{T}}(\bar{b}_{t+1} \mid \bar{b}_t, c_t)$ 
only depends on $\bar{b}_t$ and $\bar{b}_{t+1}$ through their assignments to the propositions in $\texttt{UConds}$ and $\texttt{UEffs}$ respectively.

In the first model learning iteration, Algorithm~\ref{alg:model_learning} initializes a count map $N$ where $N[\Psi_\text{pre}, c]$ is the number of times model learning has simulated controller $c$ from a belief state $b$ consistent with UCond assignment $\Psi_\text{pre}$. It also initializes map $D$ where $D[\Psi_\text{pre}, c, \Psi_\text{eff}]$ is the number of simulations in which the belief state which arose after simulating $c$ induced assignment $\Psi_\text{eff}$ to the $\texttt{UEff}$ propositions for the operator corresponding to $c$.
The algorithm also initializes an abstract to concrete belief map $P_\mathcal{B}$. (Lines~\ref{line:start_counts}-\ref{line:end_counts}.)
Inside of a model learning loop, the algorithm performs Bayes optimistic determinized planning using the Fast Downward planner with state-dependent action costs (SDAC~\cite{sdac}) derived from the partially learned model according Equation~\ref{equation:bayes_cost}. 
The resulting plans take the form of a sequence of triples $(\bar{b}_t, \texttt{op}, \bar{b}_{t+1})$ specifying that controller $\texttt{op}.c$ executed in abstract belief state $\bar{b}_t$ transitions to $\bar{b}_{t+1}$ (Line~\ref{line:plan}).
These triples are filtered to those where the abstract belief has known corresponding concrete beliefs in $P_\mathcal{B}$ (Line~\ref{line:filter}).
A subset of the remaining triples are then chosen for simulated outcome sampling; each chosen triple $(\bar{b}_1, \texttt{op}, \bar{b}_2)$ is selected if the Beta distribution describing partial knowledge about $\overline{\mathcal{T}}(\bar{b}_2 \mid \bar{b}_1, \texttt{op}.c)$ has maximal entropy $H$ among the available options (Line~\ref{line:entropy}).
After simulating $\texttt{op}.c$ from some belief state $b_1$ consistent with $\bar{b}_1$, and producing concrete belief state $b_2$, the algorithm computes the $\Psi_\text{pre}$ corresponding to $\bar{b}_1$ and the $\Psi_\text{eff}$ corresponding to $b_2$, and updates the counts in $N$ and $D$ (Lines~\ref{line:simulate}-\ref{line:update_counts_end}).
At the end of model learning, $N$ and $D$ are compiled into a transition model $\hat{\mathcal{T}}$ on the subset $\mathcal{B}_\text{sparse} \subseteq \mathcal{B}$ of abstract belief states reachable from $b_0$ by applying sequences of operators explored in model learning (Line~\ref{line:compile}). See Appendix~\ref{app:model_compilation} for details.


\subsection{Progressive widening} 
\label{sec:progressive_widening}

Per Section~\ref{sec:example_operator}, operators and controllers are derived by binding operator schemata to assignments of objects and continuous parameters.
In Algorithm~\ref{alg:model_learning}, we assume the set of continuous inputs are fixed and prespecified. The algorithm can be extended by using progressive widening to gradually expand the operator set $\mathbb{O}$, adding in new operator instances corresponding to applications of controller schemata bound to new continuous parameters drawn from a stream of sampled values (see Appendix~\ref{app:samplers}).

Because this effectively increases the state and action space of the abstract MDP, care must be taken to expand this set gradually so that the expansion of the MDP does not outpace the optimistic exploration.
To achieve this, we use a progressive widening criteria typically used in hybrid discrete-continuous search problems~\cite{pomcpow}.
Our full TAMPURA implementation incorporates progressive widening by adding a line before Line~\ref{line:plan} in Algorithm~\ref{alg:model_learning} to add elements to $\mathbb{O}$.

Such widening increases continuous action input samples based on the number of times a ground operator has been visited, maintaining the following relationship during model learning for each controller simulation from belief $b$:
\begin{equation}
k \cdot \sum^{O}_{\texttt{op}}N[\Psi_\text{pre}, \texttt{op}]^\alpha \geq \lvert\{\texttt{op}' \in \mathbb{O} : \bar{b}[\texttt{op}'.\texttt{Pre}]\}\rvert.
\end{equation}
where $\alpha<1$ and coefficient $k$ are hyperparameters. In words, the branching factor of a lifted operator expands as a function of the number of times a lifted operator has been sampled.

\subsection{Learning \texttt{UConds} from controller feedback.} \label{app:ucond_learning}
There are many cases where it is not obvious ahead of time what belief state propositions $\psi \in \Psi_\mathcal{B}$ affect the outcome distribution of a controller, making it difficult to construct an appropriate $\texttt{UCond}$ set until simulations are run and it becomes evident what aspects of the environment are relevant to the controller outcome.
For instance, simulators often know when a controller failed due to the robot colliding with a particular object, and can indicate that a proposition describing the position of this object ought to be added to the $\texttt{UCond}$ set.
Our full TAMPURA implementation allows controller simulation (Alg.~\ref{alg:model_learning}, Line~\ref{line:simulate}) to additionally return a set of propositions $\texttt{UCond+}$ which TAMPURA immediately adds to the $\texttt{UCond}$ set of the operator being simulated.
This modification does not increase the ability of TAMPURA to find correct plans in the limit (as one could conservatively start with overly large $\texttt{UCond}$ sets), but can greatly increase the algorithm's efficiency.

\begin{table*}[t]
    \centering
\begin{tabular}{l l c c c c c c}
\toprule
Model Learning & Decision Making & A & B & C & D & E-MF & E-M \\
\midrule
\rowcolor{gray!25}
Bayes Optimistic & LAO* &
\(\textbf{0.87} \pm \textbf{0.01}\) &
\(\textbf{0.66} \pm \textbf{0.07}\) &
\(\textbf{0.63} \pm \textbf{0.07}\) &
\(\textbf{0.52} \pm \textbf{0.11}\) &
\(\textbf{0.95} \pm \textbf{0.00}\) &
\(\textbf{0.81} \pm \textbf{0.02}\) \\
Bayes Optimistic & MLO &
\(0.66 \pm 0.09\) &
\(\textbf{0.65} \pm \textbf{0.07}\) &
\(0.27 \pm 0.10\) &
\(0.29 \pm 0.10\) &
\(\textbf{0.95} \pm \textbf{0.00}\) &
\(0.41 \pm 0.09\) \\
Bayes Optimistic & WAO &
\(\textbf{0.78} \pm \textbf{0.06}\) &
\(\textbf{0.70} \pm \textbf{0.07}\) &
\(0.32 \pm 0.10\) &
\(0.24 \pm 0.09\) &
\(\textbf{0.95} \pm \textbf{0.00}\) &
\(0.56 \pm 0.08\) \\
$\epsilon$-greedy & LAO* &
\(0.69 \pm 0.08\) &
\(0.58 \pm 0.07\) &
\(0.45 \pm 0.10\) &
\(\textbf{0.42} \pm \textbf{0.10}\) &
\(\textbf{0.93} \pm \textbf{0.00}\) &
\(\textbf{0.74} \pm \textbf{0.06}\) \\
None & LAO* &
\(0.00 \pm 0.00\) &
\(0.13 \pm 0.04\) &
\(0.20 \pm 0.09\) &
\(0.34 \pm 0.09\) &
\(\textbf{0.95} \pm \textbf{0.00}\) &
\(0.00 \pm 0.00\) \\
Q-Learning & Q-Learning & 
\(0.42 \pm 0.08\) &
\(0.00 \pm 0.00\) &
\(0.20 \pm 0.08\) &
\(0.34 \pm 0.10\) &
\(\textbf{0.93} \pm \textbf{0.04}\) &
\(0.00 \pm 0.00\) \\
MCTS & MCTS &
\(0.00 \pm 0.00\) &
\(0.00 \pm 0.00\) &
\(0.04 \pm 0.05\) &
\(0.24 \pm 0.09\) &
\(\textbf{0.92} \pm \textbf{0.01}\) &
\(0.03 \pm 0.03\) \\
DQN & DQN &
\(0.00 \pm 0.00\) &
\(0.00 \pm 0.00\) &
\(0.00 \pm 0.00\) &
\(\textbf{0.47} \pm \textbf{0.09}\) &
\(0.55 \pm 0.10\) &
\(0.00 \pm 0.00\) \\
\bottomrule
\end{tabular}

    \caption{
    Average and standard error of discounted return \((\gamma=0.98)\) for various model learning and decision-making strategies on the tasks in Figure~\ref{fig:tasks}. 
    Our TAMPURA algorithm is in the top row. 
    We bold all scores within a 75\% confidence interval (N=20) of the top performing approach for each task. 
    Solution times for each method and environment are reported in the Appendix; all solvers required comparable CPU time of 20-200 seconds, depending on the task (See Table~\ref{tab:planning_times}).
    }
    \label{tab:results}
    \vspace{-2mm}
\end{table*}

\section{Simulated Experiments \& Analysis}
\label{sec:real_experiments}
We applied TAMPURA to five simulated and two real-world robotics problems, illustrated in Figure~\ref{fig:tasks} and Figure~\ref{fig:teaser}, respectively.  In our simulated experiments, we compared the performance across this range of tasks to Monte Carlo tree search and reinforcement learning baselines, as well as to belief-space task and motion planning algorithms without efficient model-learning and without uncertainty-aware planning (Table~\ref{tab:results}). This section provides a brief overview of the simulated environments and baselines, with more details in Appendix~\ref{appx:tasks} and~\ref{app:experiment_details}.
All simulated robot experiments are performed in the pybullet physics simulation~\cite{pybullet}. 
Grasps are sampled using the mesh-based EMA grasp sampler proposed in~\cite{m0m}, inverse kinematics and motion planning are performed with tools from the 
pybullet planning library~\cite{garrett2018pybullet}. 

\subsection{Simulated Domains}
\label{sec:simulated_domains}
The $\textsc{Class Uncertainty}$ (A) task requires a robot to place all objects of a certain class in a bowl, despite noisy classifications (as occur when using detectors like MaskRCNN~\cite{maskrcnn}).  The $\textsc{Pose Uncertainty}$ (B) task requires the robot to stack objects with local pose uncertainty, as arises when using standard pose estimation techniques from RGB-Depth video.  The $\textsc{Partial Observability}$ (C) task requires the robot to find and pick up a hidden object in the scene. The $\textsc{Physical Uncertainty}$ (D) requires the robot to hit a puck with unknown friction parameters to a goal region.  The $\textsc{SLAM Uncertainty}$ (E-M) task is a 2D version of a mobile manipulation task, in which a robot must bring yellow blocks in the environment to a goal region, with ego-pose uncertainty increasing over time, except when the robot visits a blue localization beacon (similarly to mobile robots using AR tags to localize).  Manipulation-free SLAM variant (E-MF) just requires the agent to move to a goal region without interacting with blocks; this only requires 1-2 controller executions and was used to verify correctness of the baselines' implementations.
This suite of tasks tests the planner's competence in a range of scenarios, including planning with risk awareness, planning to gather information.

\subsection{Baselines}

\begin{figure*}[htbp]
    \centering
    \includegraphics[width=\linewidth]{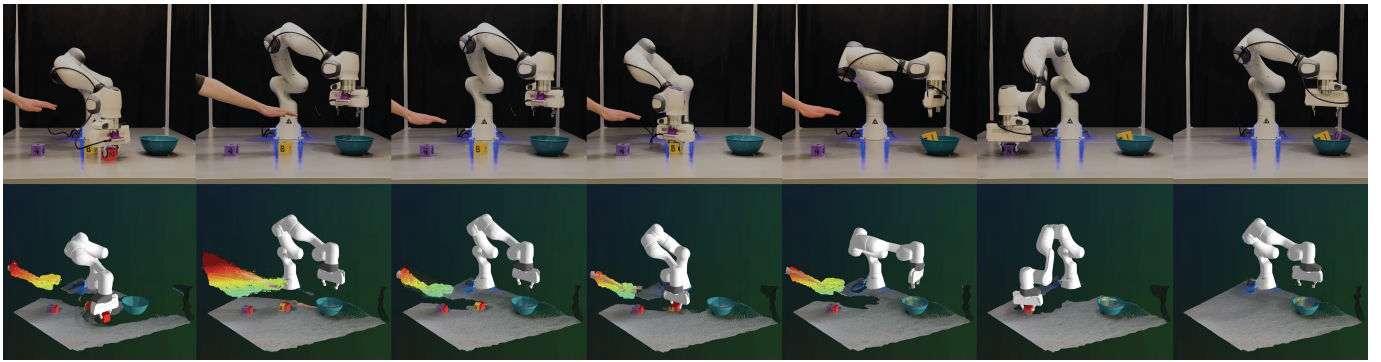}
    \caption{
    TAMPURA moving cubes into a bowl without hitting a human in the workspace.  Top row: images of robot execution.  Bottom row: the robot's belief about object poses and the probabilistic occupancy grid describing the human in the workspace.
    }
    \label{fig:real_experiments_hri}
    \vspace{-4mm}
\end{figure*}


In our simulated experiments (Table~\ref{tab:results}), we compare to many baselines from across the literature on sample-based POMDPs, reinforcement learning, and belief-space TAMP. 
Because our approach relies on closed-loop belief-space controllers, we are unable to fairly compare to point-based POMDP solvers that plan in the state space~\cite{pomcp, despot}. 
Instead we elect to compare to a variant of these planners that performs the MCTS in belief space~\cite{pomcpow, curtis2022taskdirected}.
We also compare TAMPURA to ablations resembling the limitations of previous belief-space TAMP methods.
TAMPURA uses the proposed Bayes optimistic model learning strategy, and the LAO*~\cite{lao} probabilistic planner to solve the resulting BSSP problem. 
We compare to other non-probabilistic decision-making strategies commonly used in belief-space TAMP such as weighted all outcomes (WAO) ~\cite{SSReplan} and maximum likelihood observation (MLO) determinized planning~\cite{bhpn, rohan_belief_tamp}. 
Additionally, we compare to different model learning strategies including epsilon-greedy exploration described in Section~\ref{sec:solution_guided} and contingent belief-space TAMP~\cite{contingent_tamp}, which corresponds to performing no model learning and positing an effective uniform distribution over possible observations. 

Our experiments show that Bayes optimistic model learning with full probabilistic decision making is best of these methods across this set of tasks. 
While determinized planning in belief space is sufficient for some domains like $\textsc{Pose Uncertainty}$ where most failures can be recovered from and the observational branching factor is binary, it performs poorly in domains with irreversible outcomes and higher observational branching factors.
Our experiments verify that relative to Bayesian Optimistic model learning, $\epsilon$-greedy frequently falls into local minima that it struggles to escape through random exploration in the action space.
We observed that contingent planning frequently proposes actions that are geometrically or kinematically implausible, or inefficient because it does not consider the probabilities associated with different belief-space outcomes. 
For example, in the $\textsc{Partial Observability}$ domain, we saw the robot look behind and pick up objects with equal probability rather than prioritizing large occluders.
Finally, MCTS and DQN performed poorly in most domains because they do not use high-level symbolic planning to guide their search. Without dense reward feedback, direct search in the action space is not sample efficient. 
The exception to this (other than SLAM-MF, used to verify  implementation correctness) is $\textsc{Physical Uncertainty}$, where the time horizon is short, with optimal plans only requiring 1-4 controller executions.

\section{Real-world Implementation}

We implemented TAMPURA on a Franka robot arm to solve two tasks involving partial observability and safety with human interaction. 
Our robot experiments use Realsense D415 RGB-D cameras with known intrinsics and extrinsics.
We use Bayes3D perception framework for probabilistic pose inference~\cite{gothoskar2023bayes3d}.
In our experiments we used objects with known mesh object models, but Bayes3D also supports few-shot online learning of object models.
See the supplementary material for videos of successful completions under various initializations of these tasks.

\subsection{Searching for Objects in Clutter}

This task is the real-world counterpart to the $\textsc{Partial Observability}$ simulated experiment.
In this task, the robot is equipped with a single RGBD camera mounted to the gripper, and must find and pick up a small cube hidden in the environment.
This requires looking around the environment, and potentially moving other objects out of the way to make room to see and grasp the cube.
Using Bayes3D's capacity to not only estimate poses of visible objects, but represent full posterior distributions over the latent scene given RGBD images, TAMPURA can characterize the probability that an unseen object is hidden behind each visible object.
We experimented with various object sets and arrangements, and observed qualitatively sensible plans.
For instance, TAMPURA moved larger objects with a larger probability of hiding the cube before moving smaller objects aside.
The primary failure modes were (1) failure in perception (due, we believe, to improperly calibrated hard-coded camera poses), and (2) issues with tension in the unmodelled cord connected to the camera. 
Planning sometimes failed due to insufficient grasp and camera perspective sampling, which could be resolved by increasing maximum number of samples.

\subsection{Safety in Human-Robot Interaction}
In this task, several cubes and a bowl are placed on a table.  The robot's task is to move these cubes into the bowl without colliding with a human's hand moving around in the workspace.  The robot's belief states consist of a posterior over static object poses returned by Bayes3D, and a probabilistic 3D occupancy grid representing knowledge about dynamic elements in the scene (namely, the human).  We update the occupancy belief probabilities over time as follows.
Let \( P(t, x, y, z) \) be the probability that the voxel at coordinates \((x, y, z)\) was occupied at time \(t\). 
The updated probability \( P(t + \Delta t, x, y, z) \) at time \(t + \Delta t\) is given by

\begin{equation}
P(t + \Delta t, x, y, z) = P(t, x, y, z) \times \gamma^{(C\Delta t)}
\end{equation}

\noindent where \( \gamma \) is the decay rate constant, \( C \) is the decay coefficient, and \( \Delta t \) is the time step.
At each time $t$, the current frame of RGBD video was processed to obtain a point cloud, and each voxel occupied by a point had its occupancy probability reset to 1.
A visualization of this grid can be seen in Figure~\ref{fig:teaser}.
Given the current probabilistic occupancy grid, generated from point cloud data from RGB-Depth cameras, we approximate a motion planning path with gripper interpolation and calculate collision probabilities by integrating grid cell occupancy probabilities along the trajectory. For details, see Appendix~\ref{appx:hri}.
The resulting planner is able to make high-level decisions about safety and human avoidance. 
The supplemental video contains examples of the planner picking objects in an order that has the lowest probability of collision and waiting for the human to clear from the workspace before reaching for objects.

\section{Discussion}
\label{sec:discussion}

\begin{figure*}[htbp]
    \centering
    \includegraphics[width=\linewidth]{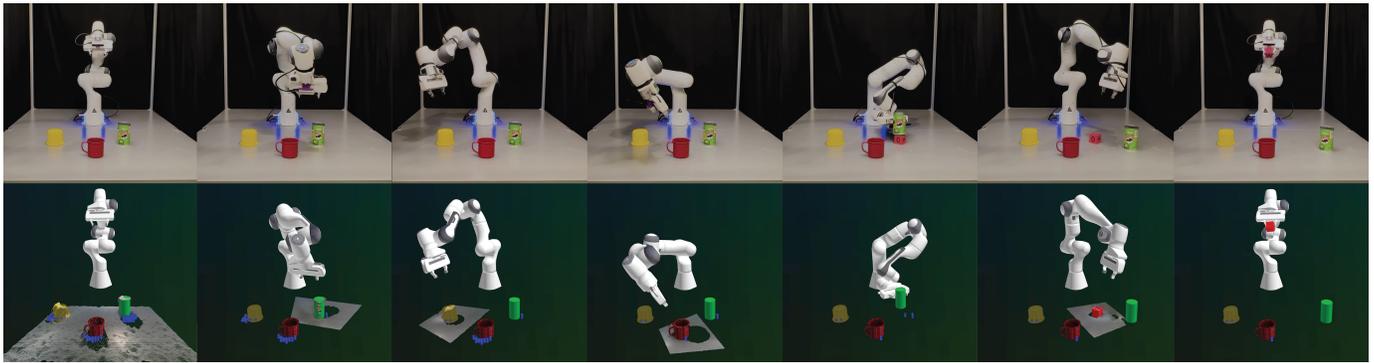}
    \caption{
    TAMPURA searching a workspace to find and pick up a cube, looking around and moving objects to find it.  Top: images of robot execution.  Bottom: 
    the robot's belief about the location of the target object over time. 
    Each blue point in the robot's belief visualization is the centroid of a possible object location in the posterior returned by Bayes3D. Since the object models are known, the robot knows that the target object could be under the green cup or yellow cups with low probability. Because the yellow cup is too large to be grasped, the robot looks under the green cup after ruling out other possible locations.
    }
    \label{fig:real_experiments_po}
\end{figure*}

The TAMP framework enables zero-shot generalization to novel objects, scenes, configurations, and tasks, but typically relies on assumptions of full-observability and deterministic outcomes. In this work, we presented a TAMP method capable of reasoning about uncertainty and risk at both the task and motion planning levels to act efficiently in arbitrary partially observable environments.
We demonstrated that resulting planner produces efficient risk and uncertainty aware plans across a range of real and simulated robotic tasks.
Our approach enables long-horizon robot planning without precise descriptions of the effects of low-level controllers, lending itself to arbitrary learned or designed controllers. 

Despite these novelties, TAMPURA, and TAMP in general, have several limitations. First, planning time can be burdensome when used in the context of real-time systems. Recent developments in faster motion planning~\cite{thomason2024vamp} or GPU-based parallel simulation~\cite{makoviychuk2021isaac} could ease this burden. Second, this planner requires user-provided belief representations, abstractions, and belief updating functionality. Designing each of these components can take considerable engineering effort, and may require expert knowledge in perception, inference, planning, and control. We believe extending this framework to handle learned abstractions and probabilistic models is an important direction for future research.

\section{Acknowledgements}
We gratefully acknowledge support from NSF grant 2214177; from AFOSR grant FA9550-22-1-0249; from ONR MURI grant N00014-22-1-2740; from ARO grant W911NF-23-1-0034; from MIT-IBM Watson Lab; from the MIT Quest for Intelligence; from DARPA, under the DARPA Machine Common Sense program (Award ID: 030523-00001) and JUMP (CoCoSys, Prime Contract No. 2023-JU-3131) program; from the Boston Dynamics Artificial Intelligence Institute; unrestricted gifts from Google; as well as philanthropic gifts from an anonymous donor and the Siegel Family Foundation. Aidan Curtis is supported by the NSF GRFP fellowship.

\bibliography{references}

\clearpage

\appendix

\subsection{Code release}
Our TAMPURA implementation, as well as full implementations of our simulated experiments (including the environments, controllers, and stream specifications), will be released at a \href{https://aidan-curtis.github.io/tampura.github.io/}{\color{blue}{public repository}}.

\subsection{Compiling simulation outcome counts into the sparse abstract MDP} \label{app:model_compilation}

Line~\ref{line:compile} of Algorithm~\ref{alg:model_learning} compiles the dictionaries $N$ and $D$ storing simulation counts into a learned transition distribution $\hat{\mathcal{T}}$ on a subset $\mathcal{B}_{\text{sparse}} \subseteq \overline{\mathcal{B}}$ of the set of all abstract beliefs.  
These components, $\mathcal{B}_\text{sparse}$ and $\hat{\mathcal{T}}$, define an MDP (which we refer to throughout as the ``sparse MDP'') that is passed into the LAO* probabilistic planner in Algorithm~\ref{alg:tampura}, for uncertainty and risk aware planning.  
We now elaborate on how $\mathcal{B}_\text{sparse}$ and $\hat{\mathcal{T}}$ are constructed.

The set $\mathcal{B}_{\text{sparse}}$ consists of all abstract belief states reachable from the current belief state $b_0$, by applying the guaranteed effects ($\texttt{Effs}$) of operators visited during model learning, and applying assignments of uncertain effects ($\texttt{UEffs}$) present in at least one simulation from model learning.
The key set of $D$ consists of values of the form $(\Psi_\text{pre}, c, \Psi_\text{eff})$, where $\Psi_\text{pre}$ is an assignment to the $\texttt{UCond}$ set of the operator corresponding to controller $c$, and $\Psi_\text{eff}$ is an assignment to its $\texttt{UEff}$s.
$D$ thereby stores the set of all operators which were simulated during model learning (as each controller $c$ corresponds to a particular operator), and all UEff assignments produced in model learning simulations.

The transition probabilities $\hat{\mathcal{T}}$ are as follows.  Consider any operator $\texttt{op} \in \mathbb{O}$, any abstract belief state $\bar{b}$, and any assignment $\Psi_\text{eff}$ to $\texttt{op}.\texttt{UEffs}$.  Let $\bar{b}'$ is the abstract belief state obtained by beginning in $\bar{b}$ and applying each effect in $\texttt{op}.\texttt{Effs}$, as well as each effect in $\texttt{op}.\texttt{UEffs}$ marked as true in $\Psi_\text{eff}$.  Let $\Psi_\text{pre}$ denote the assignment to $\texttt{op}.\texttt{UConds}$ in $\bar{b}$.  Then the transition probability is the fraction of simulations of $\texttt{op}.c$ run in model learning from $\Psi_\text{pre}$ that resulted in assignment $\Psi_\text{eff}$:
$$
\hat{\mathcal{T}}(\bar{b}' \mid \bar{b}, \texttt{op}.c) := \frac{D[\Psi_\text{pre}, \texttt{op}.c, \Psi_\text{eff}]}{N[\Psi_\text{pre}, \texttt{op}.c]}
$$

For $(\bar{b}, \texttt{op}, \bar{b}')$ where the resulting pair $(\Psi_\text{pre}, \texttt{op})$ was explore during model learning, but never produced UEffs matching $\bar{b}'$, $\hat{\mathcal{T}}(\bar{b}' \mid \bar{b}, \texttt{op}.c) := 0$.  In the case that the pair $(\Psi_\text{pre}, \texttt{op})$ was never explored during model learning, we do not even add an entry for $(\bar{b}, \texttt{op}, \bar{b}')$ to the data structure representing $\hat{\mathcal{T}}$.  The fact that $\hat{\mathcal{T}}$ does not contain any entries beginning with $(\bar{b}, \texttt{op})$ represents to the probabilistic planner that in belief state $\bar{b}$, operator $\texttt{op}$ cannot be appliex and should not even be considered (as it was never visited during model learning in an abstract belief state with UConds matching $\bar{b}$).  This is important to performant planning by LAO*, as it ensures that there are only a relatively small number of operators applicable from each abstract belief space $\bar{b}$.  It is in this sense that the MDP given to LAO* is sparse; we believe this sparsity plays a key role in the tractability of solving the MDP given to LAO*.

In many cases, the set $\mathcal{B}_{\text{sparse}}$ can be constructed explicitly by initializing $\mathcal{B}_\text{sparse}$ the set $\{\bar{b}_0\}$ just containing the initial abstract belief state, and then iteratively adding all abstract belief states that would result from applications of explored operators to the states currently in $\mathcal{B}_\text{sparse}$.  (In fact, we implemented this and used it to produce the results in Figure~\ref{fig:mdp_experiments}.
These results value iteration rather than LAO* to solve the sparse MDP, to ensure the comparison targeted the quality of the learned transition model without effects related to the interaction with an approximate MDP solver like LAO*.
Value iteration requires explicit representation of the MDP state space $\mathcal{B}_\text{sparse}$.)
However, the full TAMPURA implementation never explicitly constructs $\mathcal{B}_\text{sparse}$.
Instead, it gives LAO* the sparse MDP in the form of data structures which, for any $\bar{b} \in \mathcal{B}_\text{sparse}$, can list the operators which can be applied in $\bar{b}$, and the distribution over possible outcome abstract belief states $\bar{b}'$ induced by applying each operator.

It is the sparsity of the action branching factor that makes the sparse MDP tractable solve, not the small size of the state space.
(Indeed, $\mathcal{B}_\text{sparse}$ can be large enough we found it desirable not to have to construct it explicitly.)
The ability to learn a transition model on a relatively large set of abstract belief states, but also produce efficient probabilistic plans in the resluting MDP due to its action sparsity, is a key feature of TAMPURA's approach.
(One benefit of having $\mathcal{B}_\text{sparse}$ cover more states is that it decreases the frequency of replanning.)
The ability to learn a transition model on all of $\mathcal{B}_\text{sparse}$ derives from learning probability tables from $(\texttt{UConds}, \texttt{op})$ pairs to distributions over $\texttt{UEffs}$, rather than directly learning transition probabilities from each pair $(\bar{b}, \texttt{op})$ to a distribution on resulting abstract belief states.
(Each $\texttt{UCond}$ and $\texttt{UEff}$ assignment is consistent with many abstract belief states, so this learning representation is much more efficient.)




\subsection{Extended Task Descriptions} \label{appx:tasks}

\subsubsection{$\textsc{Class Uncertainty}$}
This task considers a robot arm mounted to a table with a set of 2 to 10 objects placed in front of it, with at least one bowl in the scene. 
The robot must place all objects of a certain class within the bowl without dropping any objects. 
We add classification noise to ground truth labels to mimic the confidence scores typically returned by object detection networks like MaskRCNN~\cite{maskrcnn}. 
Object grasps have an unknown probability of success, which can be determined through simulations during planning. 
The agent can become more certain about an object category by inspecting the object more closely with a wrist mounted camera. 
A reasonable strategy is to closely inspect objects that are likely members of the target class and stably grasp and place them in the bowl. The planner has access to the following controllers: $\texttt{Pick(?o ?g ?r)}$, $\texttt{Drop(?o ?g ?r)}$, $\texttt{Inspect(?o)}$, for objects $o$, grasps $g$, and regions on the table $r$.

\showpddl{
\begin{lstlisting}[style=PDDLStyle]
(:action pick
 :parameters (?o - object ?g - grasp ?r - region)
 :precondition (and (not (Holding)) (ObjectGrasp ?o ?g) (not (Broken ?o)) (On ?o ?r))
 :effects (and (not (Holding ?o)) (On ?o ?r))
 :uconds (and (IsClass ?o @fruit))
 :ueffects (and (AtGrasp ?o ?g) (maybe (Broken ?o)))

(:action drop
 :parameters (?o - object ?g - grasp ?r - region)
 :precondition (and (IsBowl ?r) (not (Broken ?o)) (AtGrasp ?o ?g))
 :effects (and (not (AtGrasp ?o ?g)) (On ?o ?r))

(:action inspect
 :parameters (?o - object)
 :precondition (and (not (Holding)) (PossibleClass ?o @fruit) (not (Broken ?o)))
 :ueffects (and (IsClass ?o @fruit) (PossibleClass ?o @fruit))

\end{lstlisting}
}

\subsubsection{$\textsc{Pose Uncertainty}$}
This task consists of 3 cubes placed on the surface of a table and a hook object with known pose. 
The cubes have small Gaussian pose uncertainty in the initial belief, similar to what may arise when using standard pose estimation techniques on noisy RGBD images. 
The goal is to stack the cubes with no wrist-mounted camera. 
A reasonable strategy is to use the hook to bring the objects into reach or reduce the pose uncertainty by aligning the object into the corner of the hook such that grasping and stacking success probability is higher. The planner has access to controllers $\texttt{Pick(?o, ?g)}$, $\texttt{Place(?o, ?g, ?p, ?r)}$, $\texttt{Stack(?o1, ?g, ?o2)}$, and $\texttt{Pull(?o1, ?g, ?o2)}$, for an physical objects $o, o_1, o_2$, grasps $g$, regions on the table $r$, and 3D pose $p$.  ($\texttt{Pull}$ pulls one object using another object.)

\showpddl{
\begin{lstlisting}[style=PDDLStyle]
(:action pick
 :parameters (?o - physical ?g - grasp)
 :precondition (and
                (not (exists ?o2 - physical (On ?o2 ?o)))
                (not (exists ?o2 - physical (On ?o ?o2)))
                (not (exists ?o2 - physical ?g2 - grasp (AtGrasp ?o2 ?g2)))
                (KnownPose ?o)
                (ObjectGrasp ?o ?g))
 :effect (AtGrasp ?o ?g))

(:action place
 :parameters (?o - physical ?p - pose ?g - grasp ?r - region)
 :precondition (and
                (AtGrasp ?o ?g)
                (Placement ?o ?p ?r))
 :effect (and
          (not (AtGrasp ?o ?g))
          (KnownPose ?o)))

(:action stack
 :parameters (?o1 - physical ?g1 - grasp ?o2 - physical)
 :precondition (and
                (not (Hook ?o1))
                (not (Hook ?o2))
                (not (exists ?o3 - physical (On ?o3 ?o2)))
                (AtGrasp ?o1 ?g1)
                (or (KnownPose ?o2) (Exists ?o3 - physical (On ?o2 ?o3))))
 :effect (and
          (On ?o1 ?o2)
          (not (AtGrasp ?o1 ?g1))))

(:action pull-towards
 :parameters (?o1 - physical ?g1 - grasp ?o2 - physical)
 :precondition (and
                (Hook ?o1)
                (AtGrasp ?o1 ?g1)
                (not (KnownPose ?o2)))
 :effect (KnownPose ?o2))

(:reward
 :formula (and
           (On @object1 @object2)
           (On @object2 @object3)
           ...))
\end{lstlisting}
}

\subsubsection{$\textsc{Partial Observability}$}
This task, the agent has 2 to 10 objects placed in front of it with exactly one die hidden somewhere in the scene such that it is not directly visible. 
The goal is to be holding the die without dropping any objects. 
The robot must look around the scene for the object, and may need to manipulate non-target objects under certain kinematic, geometric, or visibility constraints. The planner has access to $\texttt{Pick(?o, ?g)}$, $\texttt{Place(?o, ?g)}$, $\texttt{Look(?o, ?q)}$, and $\texttt{Move(?q)}$ controllers for this task. 

\showpddl{
\begin{lstlisting}[style=PDDLStyle]
(:axiom holding
 :parameters (?obj - physical)
 :condition (Holding ?obj))
(:action pick
 :parameters (?o - physical)
 :precondition (and (not (Moved ?o)) (not (holding)) (KnownPose ?o))
 :effects (and (not (AtHome)) (when (holding) (Holding ?o)))
 :ueffects (and (verify (Holding ?o))))
(:action place
 :parameters (?o - physical)
 :precondition (Holding ?o)
 :effects (and (Moved ?o) (not (AtHome)) (not (Holding ?o))))
(:action look
 :parameters (?o1 - physical ?o2 - physical ?q - conf)
 :precondition (and (LookingConf ?o2 ?q) (not (KnownPose ?o1)) (KnownPose ?o2) (not (holding)))
 :effects (and (not (AtHome)) (verify (KnownPose ?o1)))
 :uconds (Moved ?o2))
(:action go-home
 :parameters ()
 :precondition (and (holding) (not (AtHome)))
 :effects (AtHome))
(:axiom HoldingTarget
 :parameters (?o - physical)
 :condition (and (IsTarget ?o) (Holding ?o)))
(:reward (:exists ?o - physical (and (HoldingTarget ?o) (AtHome))))
\end{lstlisting}
}

\subsubsection{$\textsc{Physical Uncertainty}$}
This task consists of a single puck placed on a shuffleboard in front of the robot. 
The puck has a friction value drawn from a uniform distribution. The goal is to push the puck to a target region on the shuffleboard. 
The robot can attempt pushing the puck directly to the goal, but uncertainty in the puck friction leads to a low success rate. 
A more successful strategy is to push the puck around locally while maintaining reach to gather information about its friction before attempting to push to the target. The planner has access to a $\texttt{PushTo(?o, ?r)}$ controller that pushes object $o$ to a target region $r$ and $\texttt{PushDir(?o, ?d)}$ controller that pushes object $o$ with fixed velocity in a target direction $d$, both of which are implemented as velocity control in Cartesian end-effector space.

\showpddl{
\begin{lstlisting}[style=PDDLStyle]
(:action push-to
 :parameters (?o - physical ?r - region)
 :preconditions (and (In ?o @ik) (IsGoal ?r))
 :effects (and (not (In ?o @ik)) (In ?o ?r))
 :uconds (and (KnownFriction ?o))
 :ueffects (and (KnownFriction ?o)))

(:action nudge
 :parameters (?o - physical ?d - direction)
 :preconditions (In ?o @ik)
 :effects (true)
 :uconds (true)
 :ueffects (KnownFriction ?o))

(:axiom reward
 :context (In @puck @goal0)
 :implies (Increase (reward) 100))
\end{lstlisting}
}

\begin{table*}[h]
    \centering
    \begin{tabular}{l l c c c c c c}
        \toprule
        Model Learning & Decision Making &  Task A &  Task B &  Task C &  Task D &  Task E-MF &  Task E-M \\
        \midrule

       Bayes Optimistic & LAO* &
        \(28 \pm 26\) & 
        \(21 \pm 13\) & 
        \(57 \pm 38\) & 
        \(23 \pm 7\) & 
        \(31 \pm 11\) & 
        \(129 \pm 55\) \\
    
    Bayes Optimistic & MLO & 
        \(54 \pm 3\) & 
        \(38 \pm 20\) & 
        \(49 \pm 43\) & 
        \(30 \pm 41\) & 
        \(35 \pm 24\) & 
        \(90 \pm 56\) \\
    
    Bayes Optimistic & WAO &
        \(33 \pm 24\) & 
        \(29 \pm 20\) & 
        \(87 \pm 32\) & 
        \(46 \pm 41\) & 
        \(34 \pm 15\) & 
        \(100 \pm 52\) \\
    
    $\epsilon$-greedy & LAO* & 
        \(3 \pm 0\) & 
        \(40 \pm 34\) & 
        \(72 \pm 38\) & 
        \(27 \pm 15\) & 
        \(35 \pm 17\) & 
        \(110 \pm 40\) \\

    None & LAO* & 
        \(1 \pm 0\) & 
        \(15 \pm 6\) & 
        \(3 \pm 2\) & 
        \(1 \pm 0\) & 
        \(1 \pm 0\) & 
        \(1 \pm 0\) \\

    Q-Learning & Q-Learning & 
        \(10 \pm 3\) & 
        \(181 \pm 24\) & 
        \(88 \pm 54\) & 
        \(11 \pm 11\) & 
        \(72 \pm 31\) & 
        \(186 \pm 89\) \\
        
    MCTS & MCTS & 
        \(29 \pm 29\) & 
        \(60 \pm 10\) & 
        \(54 \pm 51\) & 
        \(16 \pm 7\) & 
        \(169 \pm 13\) & 
        \(207 \pm 28\) \\
        
    DQN & DQN & 
    \(17 \pm 4\) & 
    \(12 \pm 3\) & 
    \(83 \pm 34\) & 
    \(72 \pm 29\) & 
    \(28 \pm 4\) & 
    \(84 \pm 4\) \\
    \bottomrule
    
    \end{tabular}
    \caption{Average and standard deviation of per-step planning times (seconds) averaged over trials and steps within each trial. These include execution time of the selected controller in simulation.}
    \label{tab:planning_times}
\end{table*}

\subsubsection{$\textsc{SLAM Uncertainty}$}
The task is a 2D version of a mobile manipulation task, where the robot must gather yellow blocks and bring them to a green region. 
The number, location, and shape of the objects and obstacles is randomly initialized along with the starting location of the robot. 
The initial state is fully known, but the robot becomes more uncertain in its position over time due to action noise. 
The robot can localize itself at beacons similar to the way many real-world base robots use AR tags for localization. 
To verify that all baselines were implemented correctly, we also consider a manipulation-free variant of this task (SLAM-MF) requiring only 1 or 2 controller executions for success.  The goal is to enter the target region without colliding with obstacles; blocks need not be moved. The planner has access to $\texttt{MoveRegion(?r)}$, $\texttt{MoveLook(?r)}$ (which moves to region $r$ and then localizes itself by looking at a beacon0, $\texttt{MovePick(?r, ?o)}$, $\texttt{MovePlace(?r, ?o)}$, and $\texttt{MoveCorner(?r, ?c)}$ controllers. All moving controllers use a belief-space motion planner~\cite{robustrrt} except for $\texttt{MoveCorner(?r, ?c)}$ , which simply navigates to a particular the corner of a workspace.
\showpddl{
\begin{lstlisting}[style=PDDLStyle]
(:action move_pick
 :parameters (?t - target ?r - region)
 :precondition (and (NotHolding ?t) (KnownPose) (ExistsRegion ?r))
 :effects (and (forall (?r2 - region) (NotIn ?r2)) (Holding ?t) (KnownPose))
 :uconds (ExistsRegion ?r)
 :ueffects (and (Holding ?t) (KnownPose)))

(:action move_place
 :parameters (?t - target ?r - region)
 :precondition (and (KnownPose) (Holding ?t) (NotInCollision))
 :effects (and (forall (?r2 - region) (when (not (= ?r ?r2)) (NotIn ?r2)))
                (Not (Holding ?t)) (In ?r) (TargetIn ?t ?r) (KnownPose))
 :uconds (ExistsRegion ?r)
 :ueffects (and (In ?r) (TargetIn ?t ?r) (KnownPose)))

(:action move_to
 :parameters (?r - region)
 :precondition (and (NotIsBeacon ?r) (NotIsCorner ?r) (KnownPose) (NotInCollision))
 :effects (forall (?r2 - region) (when (not (= ?r ?r2)) (NotIn ?r2)))
 :uconds (ExistsRegion ?r)
 :ueffects (In ?r))

(:action move_look
 :parameters (?r - region)
 :precondition (and (KnownPose) (IsBeacon ?r) (NotInCollision))
 :effects (forall (?r2 - region) (when (not (= ?r ?r2)) (NotIn ?r2)))
 :uconds (ExistsRegion ?r)
 :ueffects (and (In ?r) (KnownPose)))

(:action move_corner
 :parameters (?c - region)
 :precondition (and (IsCorner ?c) (NotInCollision))
 :effects (forall (?r2 - region) (when (not (= ?c ?r2)) (NotIn ?r2)))
 :uconds (ExistsRegion ?r)
 :ueffects (and (KnownPose) (In ?c)))

(:axiom ExistsRegion
 :parameters (?r - region)
 :condition (Exists (?r - region) (In ?r)))

(:reward
 :condition (and (forall (?r - region ?t - target) 
                  (imply (IsGoal ?r) (TargetIn ?t ?r)))
                 (NotInCollision)))
                 
\end{lstlisting}
}

\subsection{Experimental Details} \label{app:experiment_details}
All experiments were run on a single Intel Xeon
Gold 6248 processor with 9 GB of memory. We report planning times for each algorithm and environment combination in table~\ref{tab:planning_times}. It is important to note that we did not optimize for planning time, and all of these algorithms run in an anytime fashion, meaning that planning can be terminated earlier with lower success rates. 

We now provide several more details about the two tasks we performed using TAMPURA on the real robot.

\subsubsection{Object finding}
The robot has access to a number of controllers that it could use to find and hold a small cube. A $\texttt{Pick(?o, ?g)}$ controller will grasp an object $o$ with grasp $g$, if the variance of the object pose is below some threshold. 
A $\texttt{Place(?o, ?g)}$ controller places an object $o$ held at grasp $g$ assuming the probability of collision of that placement is below some threshold. Lastly, a $\texttt{Look(?q)}$ controller moves the robot arm to a particular joint configuration $q$ and captures an RGBD image.

\subsubsection{HRI} \label{appx:hri}

The collision probability for each trajectory segment used in the motion model is determined as follows:

\begin{equation}
P_{\text{collision}} = 1 - \prod_{t=1}^{T} \prod_{i=1}^{n_t} (1 - P(t, x_i, y_i, z_i))
\end{equation}

Here, \( T \) represents the total number of time steps in the trajectory, \( n_t \) is the number of cells encountered at time step \( t \), and \( (x_i, y_i, z_i) \) are the coordinates of the \( i \)-th cell at time \( t \) along the trajectory.

The robot has access to $\texttt{Pick(?o, ?g)}$, $\texttt{Place(?o, ?g)}$ and $\texttt{Wait()}$ controllers, for objects $o$ and grasps $g$.



\showpddl{
\begin{lstlisting}[style=PDDLStyle]
(:action pick
 :parameters (?o - object ?p - pose ?g - grasp ?q - conf ?t1 - time ?t2 - time)
 :precondition (and 
                (not (exists (?o2 - object ?g2 - grasp) (AtGrasp ?o2 ?g2)))
                (AtTime ?t1)
                (NextTime ?t1 ?t2)
                (IKSol ?o ?p ?g ?q)
                (AtPose ?o ?p)
                (not (Collision)))
 :effects (and
           (AtGrasp ?o ?g)
           (not (AtPose ?o ?p))
           (AtTime ?t2)
           (not (AtTime ?t1)))
 :ueffects (and (Collision))
)

(:action place
 :parameters (?o - object ?p - pose ?g - grasp ?q - conf ?r - region ?t1 - time ?t2 - time)
 :precondition (and
                (AtGrasp ?o ?g)
                (AtTime ?t1)
                (NextTime ?t1 ?t2)
                (IKSol ?o ?p ?g ?q)
                (Placement ?o ?p ?r)
                (not (Collision)))
 :effects (and
           (AtPose ?o ?p)
           (not (AtGrasp ?o ?g))
           (AtTime ?t2)
           (not (AtTime ?t1)))
 :ueffects (and (Collision))
)

(:action wait
 :parameters (?t1 - time ?t2 - time)
 :precondition (and
                (AtTime ?t1)
                (NextTime ?t1 ?t2)
                (not (Collision)))
 :effects (and
           (AtTime ?t2)
           (not (AtTime ?t1)))
)
\end{lstlisting}
}

\subsection{Additional Baseline Details} 
\label{app:baseline_details}

\subsubsection{DQN}
\label{app:dqn_details}
In our DQN baseline, we use the CleanRL~\cite{huang2022cleanrl} implementation. The state space of DQN is a vectorized version of the belief. To vectorize beliefs, we flatten all continuous properties of the belief, and one-hot encode all discrete properties. The action space is a pre-discretized version of the original continuous action space. We perform this discretization by sampling three possible continuous samples from each sampler described in Appendix~\ref{app:samplers}. To make for a fair comparison with other methods, we limit the DQN training to 1000 simulator samples per action. At each simulator step after sufficient data exists for a single batch, we update the network. We find that after approximately 1000 such network updates, the loss converges. Data is retained in the buffer across execution steps.

We chose to implement DQN in an online fashion (i.e. simulations performed from the initial belief state) instead of in an offline fashion, which is the typical application of RL. In an offline setting, the RL agent would be trained on a distribution of possible environments. We did not attempt to compare to RL in this way because we are interested in testing zero-shot performance on novel problems that are out of distribution for an RL agent trained on a particular distribution of tasks. Testing out-of-distribution generalization of an RL agent trained on offline data would require a separate experimental setup, and is outside of the scope of this paper.

\subsubsection{MCTS}
\label{app:mcts_details}
Our MCTS implementation uses a pre-discretized action space and plans in the abstract belief space. We make selections according to the standard UCB selection criterion $UCB = \bar{X}_j + C \sqrt{\frac{2 \ln n}{n_j}}$ where $\bar{X}_j$ is the average reward obtained from node $j$,
$n$ is the total number of simulations that have been run from the parent node,
$n_j$ is the number of simulations that have been run from the child node $j$, and $C$ is the constant determining the trade-off between exploration and exploitation. We use $C=1$ in our experiments.

\subsubsection{Q-Learning}
\label{app:q_learning_details}
We perform Q learning on a pre-discretized action space and the abstract belief space. A sparse Q table is maintained due to the intractably large abstract belief space. During each simulation, the sparse Q table is updated according to the following rule

\[Q(s, a) \leftarrow Q(s, a) + \alpha \left[ R(s, a) + \gamma \max_{a'} Q(s', a') - Q(s, a) \right]\]

In our experiments we use $\alpha=0.2$, and we take a random action with an $\epsilon=0.2$. The ultimate action is selected via $\text{argmax}_a[Q(\bar{b}_0, \cdot)]$.

\subsection{Additional Experimental Statistics}

In Table~\ref{tab:additional_exp_stats} we provide some additional statistics on our simulated experiments for the TAMPURA algorithm. These statistics include the average number of abstract belief states visited during model learning on the first execution step, and the average number of controllers TAMPURA executed on the robot in order to achieve the goal.

\begin{table}[h!]
    \centering
    \begin{tabular}{lcc}
     \toprule
        Task Name & \# Visited $\bar{b}$ & \# Executed steps \\
    \midrule
        Task A & \(118.06 \pm 14.12\) & \(6.44 \pm 1.21\)\\
        Task B & \(27.81 \pm 10.53\) & \(9.69 \pm 0.68\)\\
        Task C & \(17.40 \pm 5.43\) & \(5.40 \pm 2.33\)\\
        Task D & \(5.00 \pm 0.00\) & \(3.75 \pm 1.79\)\\
        Task E-MF & \(50.70 \pm 9.26\) & \(1.20 \pm 0.51\)\\
        Task E-M & \(68.67 \pm 26.56\) & \(9.22 \pm 3.84\) \\
    \end{tabular}
    \caption{Additional Simulated Experimental Statistics}
    \label{tab:additional_exp_stats}
\end{table}

\subsection{Continuous Action Parameter Samplers} \label{app:samplers}
Continuous action parameters such as force vectors, grasps, and joint configurations are sampled from during the model learning process. Such samples are often conditioned on other elements of the action input. For example, a grasp is specific to a particular object. Likewise, an inverse kinematics solution is specific to a particular grasp, object, and object pose. These relationships are expressed as streams, as in PDDLStream~\cite{pddlstream}. Each stream is associated with generator capable of outputting an infinite stream of samples. An example of such streams for sampling object grasps, placement poses, and inverse kinematics solutions are shown below.

\begin{lstlisting}[style=PDDLStyle]
(:stream sample-grasp
 :parameters (?o - obj)
 :domain (and (Graspable ?o))
 :output (?g - grasp)
 :certified (and (Grasp ?o ?g))
\end{lstlisting}

\begin{lstlisting}[style=PDDLStyle]
(:stream sample-placement
 :parameters (?o - obj ?s - surface)
 :output (?p - pose)
 :certified (and (Pose ?o ?p) (Support ?p ?r))
\end{lstlisting}

\begin{lstlisting}[style=PDDLStyle]
(:stream sample-ik
 :parameters (?o - obj ?g - grasp ?p - pose)
 :domain (and (Grasp ?o) (Pose ?o ?p))
 :output (?q - conf)
 :certified (and (IKSol ?o ?g ?p ?q))
\end{lstlisting}

The atoms in the domain and certified of each stream are nonfluents, which means they cannot change during planning and thus do not exist in the effects of actions. These nonfluents are referenced in the preconditions of actions to enforce relationships between input objects.

\subsection{Planner Hyperparameter Details} \label{app:hyperparameter}
In all of our simulated experiments we set a maximum of $K=1000$ simulated controller executions per real execution step, kept constant across baselines. We use an MDP $\gamma$ value of 0.98 during planning and for our evaluation metric. During model learning, we query the Top-K symbolic planner with a batch size of 20 symbolic plans. For baselines using progressive widening, we use $k=3$ and $\alpha=0.2$. Lastly, we allow all methods to run for a maximum of 20 environment steps before the planner is terminated with a failure result.

\subsection{Fast Downward Planning Details}

Our algorithm uses Fast Downward~\cite{fast_downward} planner to solve deterministic planning problems during model learning. The input to Fast Downward is a problem file describing the initially true atoms in PDDL and a domain model describing the deterministic transition model in terms of a set of action schema. All-outcome determinization of the stochastic transition dynamics described in Section~\ref{sec:operators} to standard determinisic transition dynamics is done by creating a separate action for each possible outcome. Costs are then added for each outcome's estimated log probability of occurrence using state-dependent action costs ~\cite{sdac}.

We attain batches of determinized plans using a SymK~\cite{symk}, which is a Top-K planner built on Fast Downward. The batch size to query the planner with is a user-selected hyperparameter. We run FastDownward with A-star search using the Landmark Cut heuristic.

\subsection{Stationarity of the abstract belief state MDP} \label{app:stationarity}

The condition needed for $\overline{\mathcal{T}}$ to be well defined is that the that there exists a stationary probability kernel $P(b ; \bar{b})$ describing the probability that the agent is in belief state $b$, given that the abstracted version of its belief state is $\bar{b}$.  In this paper, we assume the abstractions resulting from the user-provided operators are stationary in this sense.
Verifying this soundness property and learning sound abstractions, as is sometimes done in fully observable TAMP~\cite{gentamp, loft, predicate_invention}, is a valuable direction for future work.

We now formally define the stationarity condition needed for $P(b ; \bar{b})$, and hence $\overline{\mathcal{T}}$, to be well defined.
Let $$
    \mathcal{B}_{\bar{b}} := \{b \in \mathcal{B} : \text{abs}(b) = \bar{b}\}
$$

For each operator $c$, let $\mathcal{T}(b' \mid b, c)$ denote the probability distribution on the belief state resulting from running the controller in $c$ beginning from belief $b$.

For any $t \in \mathbb{N}$ and any sequence of operators $c_1, \dots, c_t$, consider the probability distribution $P(b \mid c_1, \dots, c_t)$ over the robot's belief state after applying the sequence of controllers $c_1, \dots, c_t$:

\begin{multline}
    P(b \mid c_1, \dots, c_t) = 
    \sum_{b_{t-1} \in \mathcal{B}} \sum_{b_{t-2} \in \mathcal{B}} \dots \sum_{b_1 \in \mathcal{B}} \\
    \mathcal{T}(b \mid b_{t-1}, c_t) \mathcal{T}(b_{t-1} \mid b_{t-2}, c_{t-1}) \dots \mathcal{T}(b_1 \mid b_0, c_1) 
\end{multline}

We will define $P(b \mid \bar{b}, c_1, \dots, c_t)$ to denote the conditional probability of being in belief state $b$ after applying operator sequence $c_1, \dots, c_t$, given that the abstract belief state corresponding to $b$ is $\bar{b}$.  For each $b \notin \mathcal{B}_{\bar{b}}$, $P(b \mid \bar{b}, c_1, \dots, c_t) := 0$, and for $b \in \mathcal{B}_{\bar{b}}$,

\begin{equation}
    P(b \mid \bar{b}, c_1, \dots, c_t) := \frac{P(b \mid c_1, \dots, c_t)}{
            \sum_{b' \in \mathcal{B}_{\bar{b}}}{P(b' \mid c_1, \dots, c_t)}}
\end{equation}

The abstract belief-state MDP is well defined so long as there exists some probability kernel $P(b ; \bar{b})$ from $\overline{\mathcal{B}}$ to $\mathcal{B}$ such that for all $t \in \mathbb{N}$ and all operator sequences $c_1, \dots, c_t$,
\begin{equation}
    P(b \mid \bar{b}, c_1, \dots, c_t) = P(b ; \bar{b})
\end{equation}
That is, given an abstract belief state $\bar{b}$, the distribution over the concrete belief state corresponding to this is independent of the time elapsed in the environment and the controllers which have been executed so far.

We anticipate that in most applications of TAMPURA, the user-provided abstractions will be imperfect, and this stationarity property will not hold exactly.  That is, the distribution $P(b \mid \bar{b}, c_1, \dots, c_t)$ will vary in $t$ and $c_1, \dots, c_t$.  The TAMPURA algorithm can still be run in such cases, as TAMPURA never computes $P(b ; \bar{b})$ exactly, but instead approximates this by constructing a table of all $(b, \bar{b})$ pairs encountered in simulations it has run.  In the case where $P(b ; \bar{b})$ is not well defined, but there exists a bound on the divergence between any pair of distributions $P(b \mid \bar{b}, c_1, \dots, c_t)$ and $P(b \mid \bar{b}, c'_1, \dots, c'_{t'})$, we expect that TAMPURA can be understood as approximately solving an abstract belief state MDP whose transition function is built from any kernel $\tilde{P}(b ; \bar{b})$ with bounded divergence from all the $P(b \mid \bar{b}, c_1, \dots, c_t)$. We leave formal analysis of TAMPURA under boundedly nonstationary abstractions to future work.

\begin{table*}[h!]
\begin{tabular}{>{}l l p{0.75\textwidth}} 
\multicolumn{2}{l}{\textbf{MDP Components}} \\\\
$\mathcal{S}, \mathcal{O}, \mathcal{A}, \mathcal{Z}, \gamma$ & (\ref{sec:background}) & State space, observation space, action space, observation function, and discount factor in the original POMDP. \\
$\mathcal{M}, \mathcal{M}_b, \mathcal{M}_c$ & (\ref{sec:background}, \ref{sec:belief_state_mdp}, \ref{sec:belief_state_controller_mdp}) & The original MDP, belief-state MDP, and belief-state controller MDP, respectively.  \\
$\mathcal{T}, \mathcal{T}_b, \mathcal{T}_c$& (\ref{sec:background}, \ref{sec:belief_state_mdp}, \ref{sec:belief_state_controller_mdp}) & State transition probabilities for original POMDP, belief-state MDP, and belief-state controller MDP, respectively. \\
$r, r_b, r_c$& (\ref{sec:background}, \ref{sec:belief_state_mdp}, \ref{sec:belief_state_controller_mdp}) & Reward functions in the original POMDP, belief-state MDP, and belief-state controller MDP, respectively. \\
$\mathcal{B}$& (\ref{sec:belief_state_mdp}) & The belief space. \\
$b_0$& (\ref{sec:belief_state_mdp}) & Initial belief state in the planning problem. \\
$\mathcal{C}$& (\ref{sec:belief_state_controller_mdp}) & The space of controllers. Each controller $c \in \mathcal{C}$ is an eventually-terminating control policy that can be executed within the belief-state MDP. \\

\\ \multicolumn{2}{l}{\textbf{Abstraction}} \\\\

$\overline{\mathcal{B}}$& (\ref{sec:abstraction}) & Abstract belief space, partitioning the continuous belief space into operationally similar groups based on belief state propositions. \\
$\overline{\mathcal{T}}$& (\ref{sec:abstraction}) & Abstract transition model giving probabilities $\overline{\mathcal{T}}(\bar{b}' \mid \bar{b}, c)$ of arriving in abstract belief $\bar{b}'$ after running controller $c$ in abstract belief state $\bar{b}$. This is the transition model for the abstract belief-state MDP $\overline{\mathcal{M}_c}$. \\
$\overline{\mathcal{M}_c}$ & (\ref{sec:abstraction}) & The abstract belief-state controller MDP. \\
$\mathcal{M}_s$& (\ref{sec:abstraction}) & The sparse abstract MDP we aim to learn and solve with a probabilistic planner \\
$\mathcal{B}_{\text{sparse}}$& (\ref{sec:background}) & The sparse belief space  of the reduced mdp $\mathcal{M}_s$ that comes out of model learning \\
$\Psi_\mathcal{B}$& (\ref{sec:belief_state_propositions}) & Set of belief state propositions used to define the abstract belief states within the belief-state MDP. \\
$\text{abs}$& (\ref{sec:abstract_bs_mdp}) & A map from beliefs in $\mathcal{B}$ to abstract beliefs in $\overline{\mathcal{B}}$ \\
$\bar{b}_0$& (\ref{sec:abstract_bs_mdp}) & Initial abstract belief state, derived from the initial concrete belief state through the application of belief state propositions. \\
$G$& (\ref{sec:abstract_bs_mdp}) & The set of abstract belief states that represents the goal of our planning problem. \\
$\mathbb{O}$& (\ref{sec:operators}) & Set of operations or actions available in the planning environment. \\
\texttt{op} & (\ref{sec:operators}) & An operator (an element of $\mathbb{O}$). This is a tuple of values $\langle \texttt{Pre}, \texttt{Eff}, \texttt{UEff}, \texttt{UCond}, c \rangle$. These values are sometimes denoted $\texttt{op}.\texttt{Pre}$, $\texttt{op}.\texttt{Eff}$, $\texttt{op}.\texttt{UEff}$, $\texttt{op}.\texttt{UCond}$, and $\texttt{op}.c$. \\

\\ \multicolumn{2}{l}{\textbf{Model Learning}} \\\\

$\alpha, \beta$& (\ref{sec:bayes_optimistic_model_learning}) & Parameters for Bayesian model learning prior. \\
$\mathcal{T}_{\text{AO}}$& (\ref{sec:bayes_optimistic_model_learning}) & The all-outcome determinized version of the abstract transition model, constructed by making all possible action/outcome combinations separate actions with deterministic outcomes. \\
$J$& (\ref{sec:bayes_optimistic_model_learning}) & A model-learning hyperparameter defining the cost of an action for a determinized version of a stochastic planning problem. \\

\\ \multicolumn{2}{l}{\textbf{Algorithm Structures}} \\\\

$I, K, S$& (\ref{sec:tampura_algorithm}) & Parameters controlling runtime: $I$ is the total iterations, $K$ is the number of trajectories, $S$ is the number of simulations. \\
$s$& (\ref{sec:tampura_algorithm}) & State from past iterations of model learning. This is initialized to $\emptyset$; after the first iteration of model learning it is a 3-tuple $(N, D, P_\mathcal{B})$.\\
$N, D$& (\ref{sec:tampura_algorithm}) & Default dictionaries used for counting simulations performed during model learning. $N[\Psi_\text{pre}, c]$ is the number of times controller $c$ was simulated from a belief state consistent with precondition set $\Psi_\text{pre}$. $D[\Psi_\text{pre}, c, \Psi_\text{eff}]$ is the number of times that controller $c$ was simulated from a belief state consistent with precondition set $\Psi_\text{pre}$, and the resulting belief state was consistent with effect set $\psi_\text{eff}$. \\
$P_\mathcal{B}$& (\ref{sec:tampura_algorithm}) & Mapping from abstract beliefs to corresponding concrete beliefs. This is a DefaultDict object. The keys are abstract beliefs, and the values are lists of concrete beliefs. The default value is the empty list.\\
$\tau_k$& (\ref{sec:tampura_algorithm}) & simulated controller execution trajectories during model learning consisting of a list of low-level action, observation pairs. \\
$\vec{\Psi}_{\text{pre}}, \vec{\Psi}_{\text{eff}}$& (\ref{sec:tampura_algorithm}) & Vectors of preconditions and effects for each transition.
Each $\Psi_\text{pre}$ in $\vec{\Psi}_{\text{pre}}$ is a boolean assignment to each predicate in $\texttt{op}.\texttt{UCond}$, for some operator $\texttt{op}$.  Each $\Psi_\text{eff}$ is a boolean assignment to each predicate in $\texttt{op}.\texttt{UEff}$.
\\
$\vec{c}$& (\ref{sec:tampura_algorithm}) & List of controllers involved in transitions. \\
$\vec{s}, \vec{f}$& (\ref{sec:tampura_algorithm}) & Vectors tracking successful and failed transitions, respectively. \\
$\vec{H}$& (\ref{sec:tampura_algorithm}) & Vector of entropy values calculated to focus simulations on uncertain cases. \\
$k, \alpha$& (\ref{sec:progressive_widening}) & Hyperparameters in progressive widening, influencing the expansion rate of action space based on the sampling frequency of actions. \\

\end{tabular}
\caption{Notation reference.} \label{table:nomenclature}
\end{table*}

\end{document}